\documentclass{article}
\usepackage{natbib}
\setcitestyle{round}
\usepackage[affil-it]{authblk}
\usepackage[english]{babel}
\usepackage[letterpaper,top=2cm,bottom=2cm,left=3cm,right=3cm,marginparwidth=1.75cm]{geometry}

\usepackage{amsmath}
\DeclareMathOperator*{\argmax}{arg\,max}
\usepackage{graphicx}
\usepackage{float}
\usepackage[T1]{fontenc}
\usepackage{enumitem}
\usepackage{bm}
\usepackage{comment}
\usepackage{subcaption}
\usepackage{booktabs, dcolumn, caption}
\usepackage{skak}
\usepackage{chessfss}
\usepackage{xcolor}
\usepackage{amsfonts}
\usepackage[colorlinks=true, allcolors=blue, breaklinks=true]{hyperref}

\title{Impartial Games: A Challenge for Reinforcement Learning}

\author[1]{Bei Zhou\thanks{Corresponding author: bei.zhou@imperial.ac.uk}}
\author[2]{S{\o}ren Riis}

\affil[1]{Imperial College London, London, United Kingdom}
\affil[2]{Queen Mary University of London, London, United Kingdom}

\date{}

\begin{document}
\maketitle

\begin{abstract}
AlphaZero-style reinforcement learning (RL) algorithms have achieved superhuman performance in many complex board games such as Chess, Shogi, and Go. However, we showcase that these algorithms encounter significant and fundamental challenges when applied to impartial games, a class where players share game pieces and optimal strategy often relies on abstract mathematical principles. Specifically, we utilise the game of Nim as a concrete and illustrative case study to reveal critical limitations of AlphaZero-style and similar self-play RL algorithms. We introduce a novel conceptual framework distinguishing between champion and expert mastery to evaluate RL agent performance. Our findings reveal that while AlphaZero-style agents can achieve champion-level play on very small Nim boards, their learning progression severely degrades as the board size increases. This difficulty stems not merely from complex data distributions or noisy labels, but from a deeper representational bottleneck: the inherent struggle of generic neural networks to implicitly learn abstract, non-associative functions like parity, which are crucial for optimal play in impartial games. This limitation causes a critical breakdown in the positive feedback loop essential for self-play RL, preventing effective learning beyond rote memorisation of frequently observed states. These results align with broader concerns regarding AlphaZero-style algorithms' vulnerability to adversarial attacks, highlighting their inability to truly master all legal game states. Our work underscores that simple hyperparameter adjustments are insufficient to overcome these challenges, establishing a crucial foundation for the development of fundamentally novel algorithmic approaches, potentially involving neuro-symbolic or meta-learning paradigms, to bridge the gap towards true expert-level AI in combinatorial games.
\end{abstract}

\section{Introduction}
\label{sec:impartial_introduction}

The advent of AlphaZero and its variations has marked a significant success in machine learning, demonstrating superhuman capabilities across complex strategy games such as chess, Go, and shogi \citep{silver2017mastering,silver2018general,schrittwieser2020mastering, sadler2019game}. These self-play reinforcement learning (RL) algorithms, leveraging powerful computational resources, can achieve superhuman performance within hours. However, despite these remarkable successes, AlphaZero-style algorithms exhibit inherent limitations, particularly concerning their robustness and generalisation to all possible game states. For instance, RL agents like KataGo have been shown to be vulnerable to adversarial attacks \citep{wang2023adversarial} and perturbations \citep{lan2022alphazero}, and may even overlook optimal moves in specific game states, indicating an incomplete mastery of the game's entire legal state space \citep{wu2019accelerating}. The identified vulnerabilities and struggles with specific game structures highlight critical gaps in the generalisation capabilities and theoretical understanding of these advanced RL systems. Addressing these limitations is paramount for advancing robust and truly general artificial intelligence, especially as complex decision-making systems become more pervasive.

This paper investigates a distinct class of games called impartial games, where AlphaZero-style algorithms encounter fundamental challenges. Unlike partisan games (e.g., chess, Go), impartial games involve players sharing game pieces, often requiring the implicit learning of a parity function to determine optimal moves. The Nim game, a canonical example of an impartial game, prominently features such a requirement. According to the Sprague–Grundy theorem, every impartial game can be reduced to an equivalent Nim-heap \citep{berlekamp2001winning}, establishing Nim's central role in impartial game theory. This makes impartial games an ideal testbed for probing the generalisation boundaries of current RL paradigms.

Learning parity functions, especially from a uniform data distribution, is a known challenge for neural networks \citep{abbe2023provable, shalev2017failures, daniely2020learning, cornacchia2023mathematical, raz2018fast, shoshani2025hardness, han2025attention}. While self-play RL agents generate non-uniform game positions, potentially creating a curriculum learning environment, empirical evidence suggests that the inherent 'noisy label' problem in self-play RL significantly impedes neural networks' ability to model parity functions \citep{zhou2023exploring}. Specifically, it has been shown that a modest rate of incorrect labels can degrade parity function learning to random guessing for bitstrings of moderate length.

Building on these theoretical considerations, particularly the statistical neutrality of parity functions \citep{thornton1996parity}, we hypothesise that AlphaZero-style algorithms, without specialised modifications, will struggle to master Nim and other impartial games on sufficiently large boards. This challenge primarily stems from the difficulty neural networks face in modelling parity functions, rather than solely from algorithmic limitations. Overcoming this requires either explicit incorporation of combinatorial game theory principles or the development of more robust neural network architectures capable of discerning such complex parity-based game properties.

To explore this hypothesis and gain deeper insights into the challenges faced by RL agents in mastering Nim-like games, we developed and extensively analysed an AlphaZero-style algorithm tailored for Nim. Our investigation also involved revisiting AlphaZero's performance in chess, focusing on positions where parity-related problems are crucial for identifying optimal moves. This comprehensive exploration culminated in a detailed theoretical and practical analysis of issues related to impartial games and parity problems. We introduce two novel metrics for evaluating self-play RL agents' performance, 'champion' and 'expert' perspectives, which offer a nuanced assessment of mastery beyond conventional winning rates.

Our findings reveal that AlphaZero-style algorithms exhibit significant difficulties in effectively learning to play Nim, especially as the board size increases. Despite Nim's low inherent complexity and simple optimal strategy from both human and computational perspectives, the algorithms struggle. While the algorithm can achieve proficiency on smaller boards or towards endgames through exhaustive search, larger, non-trivial board positions frequently resulted in the policy network failing to provide valuable guidance to the Monte Carlo Tree Search (MCTS), and the value network performing no better than random guessing. These scenarios represent 'blind spots' where optimal actions are entirely overlooked \citep{lan2022alphazero, wu2019accelerating}.

In games like chess, AlphaZero's Policy-Value Neural Networks (PV-NNs) are imperfect, but this is often compensated for by the MCTS. For some Nim boards, this imperfection might not be critical, as the policy network can guide the winning player towards positions that do not demand precise evaluations. However, we contend that this advantage diminishes on sufficiently large Nim boards due to the PV-NNs' inherent inability to model parity functions, leading to two critical issues:
\begin{enumerate}[label=(\arabic*), noitemsep]
    \item The policy network is unable to identify high-quality moves better than random guessing. \label{enu:one}
    \item The value network is unable to evaluate a position better than random guessing. \label{enu:two}
\end{enumerate}
Consequently, the policy network fails to select relevant candidate moves for optimal play, and any long-term evaluations provided by the value network become unreliable. This can also be intuitively explained by the 'statistical neutrality' of parity functions in impartial games: covering a small, unknown part of an impartial game position can render the entire position's evaluation impossible, as there is zero correlation between the visible portion and the correct evaluation. This contrasts sharply with partisan games (e.g., Go, chess), where partially observed boards typically retain positive correlation with the full board's evaluation. This inherent characteristic suggests that even minimal noise or partial observability can disrupt the positive feedback loop essential for AlphaZero-style algorithms to bootstrap learning effectively.

Our primary contributions are:
\begin{itemize}[noitemsep]
    \item We identify a distinct category of strategic games (impartial games) that necessitate non-trivial modifications to existing AlphaZero-style algorithms or current RL methodologies, highlighting a significant challenge for general game AI.
    \item We demonstrate that the challenges RL algorithms face with impartial games are more profound than simple parity-related issues, as these complexities compound, significantly reducing the self-play RL algorithm's effectiveness, and restricting its proficiency in Nim to handling fewer than 10 heaps, a stark contrast to the anticipated capability with 50+ heaps.
    \item We introduce champion and expert as definitive levels of mastery for evaluating self-play RL agent performance, providing essential metrics for nuanced assessment beyond win/loss ratios.
    \item We implemented and extensively examined an AlphaZero-style algorithm for the Nim game, revealing that mere hyperparameter adjustments are largely ineffective in mitigating the identified learning challenges.
    \item We outline potential directions for expanding the AlphaZero paradigm, laying a foundation for the creation of innovative algorithmic approaches that can tackle games with strong parity-based structures.
\end{itemize}

The remainder of this paper is structured as follows: Section~\ref{sec:impartial_games} introduces the Nim game and its relationship to other impartial games via the Sprague-Grundy theorem. We also discuss the theoretical limitations of neural network architectures in learning the parity function.  In Section~\ref{sec:impartial_revisiting}, we revisit AlphaZero and Leela Chess Zero (LCZero) to analyse their strengths and subtle weaknesses, particularly regarding parity-related issues impacting their PV-NN. Section~\ref{sec:impartial_two} defines our proposed 'champion' and 'expert' mastery levels with illustrative examples. Section~\ref{sec:impartial_rl} presents an overview of the AlphaZero-style algorithms and our re-implementation\footnote{The code for the experiments in this paper is publicly available at: \url{https://github.com/sagebei/Impartial-games-RL}}. We analyse the performance of the trained algorithm from the champion and expert perspectives, respectively, and then examine the moves made on some Nim board positions. We further conducted controlled experiments to isolate whether the challenges of AlphaZero-style reinforcement learning on impartial Nim are fundamentally caused by limitations of reinforcement learning or the inherent difficulty neural networks face in learning the parity-based winning strategy required for optimal play. We conclude the paper with general remarks, conjectures and directions for further research in Section~\ref{sec:impartial_concluding}.

\section{Impartial games and learning parity}
\label{sec:impartial_games}

\subsection{Impartial games and Nim}
The class of impartial games constitutes a significant subclass of combinatorial games \citep{berlekamp2001winning,berlekamp2002winning,berlekamp2003winning,berlekamp2004winning}. This category encompasses a diverse range of games, including classic examples such as Take-and-break, Subtraction, Heap, and Poset games. It also includes mathematically intriguing games like Nim, Sprouts, Treblecross, Cutcake, Guiles, Wyt Queens, Kayles, Grundy's Game, Quarto, Cram, Chomp, Subtract a Square, and Notakto \citep{berlekamp2001winning,berlekamp2002winning}. Many of these impartial games feature multiple variants, adding to their complexity and research interest.

An impartial game is a two-player game characterised by identical sets of available moves from any given position, irrespective of whose turn it is. Players take alternating turns. A player loses if they cannot make a legal move on their turn, meaning the player who makes the last legal move wins. Critically, any position within an impartial game can be definitively classified as either a losing (P-position, previous player winning) or winning (N-position, next player winning) position, indicating whether the current player has no winning move or at least one winning move, respectively.

\begin{figure}[ht]
\centering
\includegraphics[width=0.8\textwidth]{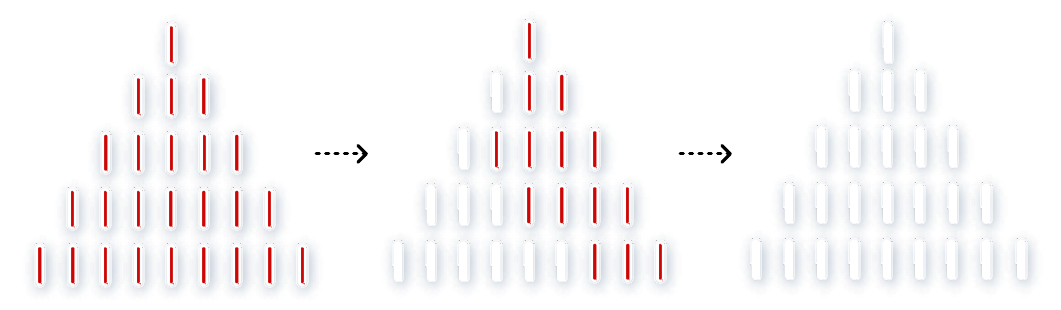}
\caption{\label{fig:nim_board} An illustration of Nim game states. The left panel shows an initial board configuration of five heaps: $[n_1,n_2,n_3,n_4,n_5]=[1,3,5,7,9]$. The middle panel depicts an intermediate board state during gameplay: $[v_1, v_2, v_3, v_4, v_5]=[1,2,4,4,3]$, resulting from players removing counters. The right panel represents the terminal state where all counters are cleared, signifying a win for the player who made the last move.}
\end{figure}

Nim serves as an important example of an impartial game and is often studied within mathematical game theory due to its elegant and analytically derived solution. Nim is played by two players who take turns removing counters from distinct heaps (or piles) arranged in a row \citep{bouton1901nim, nowakowski1998games}. On each turn, a player must remove at least one counter and may remove any number of counters, provided they are all drawn from a single heap. The objective is to be the player who removes the last counter, thereby leaving an empty board for the opponent. An initial Nim board can be represented as an array of heap sizes: $[n_1, n_2, \ldots, n_k]$, where each $n_k \in \{1, 3, \dots, 2k-1\}$ typically denotes the initial size of the $k$-th heap. The maximum number of legal moves from an initial position $[n_1, n_2, \ldots, n_k]$ is $\sum_{j=1}^k n_j$, as each move requires removing at least one counter. A game position can be represented as an array of numbers $[v_1, v_2, \ldots, v_k]$, where $v_j \leq n_j$. Although Nim is often studied in abstract terms with arbitrary configurations, RL algorithms require a fixed board size, and each self-play game begins from a predefined initial state. Figure \ref{fig:nim_board} visually illustrates an initial Nim board, an intermediate game state, and the terminal winning condition.

A property of Nim is that for any given position, it is mathematically straightforward to determine the winning player and identify all available winning moves. This determination is achieved by calculating the binary digital sum of the number of counters in each heap, a process known as the \textit{nim-sum} \citep{bouton1901nim}. This operation involves summing the heap sizes in binary without carrying over bits. The computational complexity of calculating the nim-sum has been shown to be linear in the number of heaps and logarithmic in their sizes \citep{fraenkel2004complexity}, requiring only logarithmic space memory \citep{calabro2006complexity}. Consequently, determining whether a given Nim position is a winning or losing state is computationally efficient and deterministic. This analytical solvability stands in stark contrast to the challenges faced by reinforcement learning algorithms in games lacking such perfect information or straightforward winning criteria.

Within the broader framework of combinatorial game theory, the Sprague-Grundy Theorem establishes a profound equivalence: every finite impartial game under the normal play convention (last player to move wins) is equivalent to a single-heap Nim game. This theorem assigns a "nimber" or Sprague-Grundy value, $G(s)$, to every position $s$ in an impartial game, defined recursively as:
\begin{equation}
    G(s) = \text{mex}(\{G(s') \mid s' \in N(s)\})
\end{equation}
where $N(s)$ represents the set of all states $s'$ reachable from $s$ in one legal move, and the mex (minimum excluded value) function returns the smallest non-negative integer not present in a given set \citep{beling2020pruning}. A position with Sprague-Grundy value $G(s)$ behaves identically to a Nim heap of size $G(s)$. Crucially, a position is a losing position if and only if its Sprague-Grundy value is $0$. This theorem implies that the insights gained from studying Nim's learnability extend naturally to a wide array of other impartial games.

\begin{figure}[H]
\centering
\begin{subfigure}[b]{0.4\textwidth}
\centering
\includegraphics[width=0.3\textwidth]{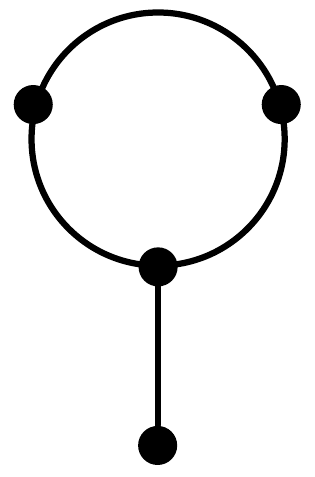}
\caption{A typical Sprouts position with a nimber value of 3.}
\label{fig:sprout_a}
\end{subfigure}
\hspace{0.01em}
\begin{subfigure}[b]{0.5\textwidth}
\centering
\includegraphics[width=0.3\textwidth]{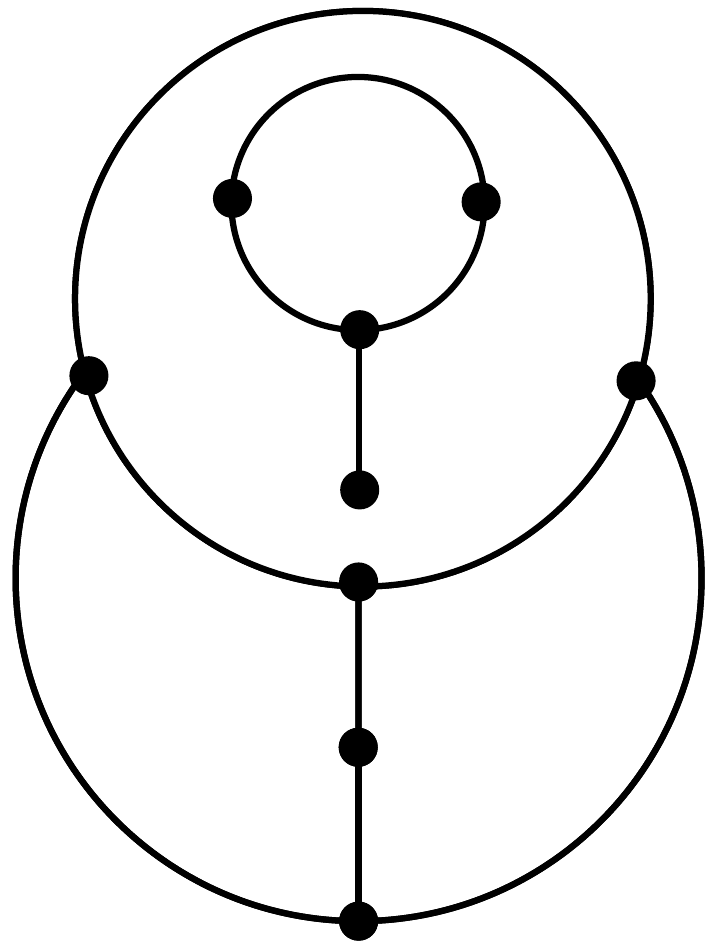}
\caption{A Sprouts position designed as an isolated gadget, also having a nimber value of 3. This gadget can be used to construct larger impartial games that behave like Nim (adapted from p. 599 in \cite{berlekamp2003winning}).}
\label{fig:sprout_b}
\end{subfigure}
\caption{\label{fig:Sprout} Mimicking Nim with Sprouts gadgets. A game of Nim played on a board $[3,3,\ldots,3]$ with $n$ heaps is equivalent to a Sprouts game constructed with $n$ copies of the isolated gadget shown in Figure \ref{fig:sprout_b}.}
\end{figure}

The theoretical underpinnings of impartial games are closely connected to the nim-sum, which in turn relies on the concept of the parity function. Thus, the parity function plays an implicit or explicit central role in the theory of impartial games, as many such games can be shown to mimic Nim or parts of Nim. To illustrate this fundamental connection, consider the impartial game Sprouts \citep{berlekamp2001winning, gardner1967mathematical}, a relatively simple game invented by John Conway and Michael Paterson. Positions in Sprouts typically yield nimber values of 0, 1, 2, or 3 \citep{berlekamp2003winning}. As depicted in Figure \ref{fig:sprout_a}, a specific Sprouts position can have a nimber value of 3. Furthermore, a modified Sprouts position (Figure \ref{fig:sprout_b}), designed as an "isolated land" or "gadget" that cannot interact with external elements, also possesses a nimber value of 3. A Sprouts starting position comprising $n$ copies of the gadget shown in Figure \ref{fig:sprout_b} effectively mimics any Nim position reachable from a starting position with $n$ heaps, each containing 3 counters.

Despite the analytical simplicity of Nim, some impartial games present significant computational challenges, as their complexity can rise to PSPACE-complete. From an algorithmic perspective, while many impartial games share the linear-time solvability of Nim, certain others, such as Node Kayles and Geography, are demonstrably harder, being PSPACE-complete \citep{schaefer1978complexity}. Moreover, even some partisan games, where move legality depends on the player, can exhibit properties akin to impartial games, such as Dawson's chess, which is effectively an impartial game in disguise \citep{berlekamp2001winning}.

Historically, algorithms for impartial games have predominantly relied on handcrafted programs incorporating principles similar to alpha-beta search, but specifically tailored for the mex operation and Sprague-Grundy values \citep{viennot2007further}. For instance, \cite{beling2020pruning} introduced a novel method for pruning search trees based on node values computed via the mex function, particularly effective for relatively "short" impartial games like Nim, Chomp, and Cram. However, this approach typically does not scale efficiently to games with large board sizes or extensive state spaces. These conventional, handcrafted programs, which leverage the explicit mathematical structure of impartial games, serve as crucial benchmarks against which the performance and generalisation capabilities of emerging reinforcement learning-based algorithms can be rigorously evaluated. The objective is to assess whether contemporary RL methods can not only match but potentially surpass the limitations of these specialised traditional approaches, particularly for larger or more complex instances of impartial games where an explicit solution might be intractable.

\subsection{Learning parity with neural networks}
\label{sec:learning_parity_nn}
The parity function is a foundational challenge that reveals the inherent limitations of various neural network architectures in learning abstract, non-associative logic, a critical skill for optimal play in impartial games. Given a bitstring $x = (x_1, x_2, \dots, x_n) \in \{0, 1\}^n$, the parity function $\mathcal{P}_n(x)$ is formally defined as:
\begin{equation}
\label{eq:parity_definition}
\mathcal{P}_n(x) = \left( \sum_{i=1}^{n} x_i \right) \pmod{2}
\end{equation}
The output is $1$ (odd parity) if the number of $1$'s in the bitstring is odd, and $0$ (even parity) otherwise.

Early work focused on constructing specialised NNs with fixed weights to solve the parity problem perfectly \citep{hohil1999solving, liu2002n, franco2001generalization, wilamowski2003solving}. However, these models cannot generalise to real-world scenarios where the underlying data distribution is unknown and weights must be learned through training.

Focusing on the challenge of learning, Recurrent Neural Networks (RNNs), including LSTMs, face significant theoretical and practical difficulties. Although RNNs are theoretically capable of simulating any Turing Machine \citep{siegelmann1995computational}, this relies on unrealistic assumptions like infinite precision and computation time. In practice, training these networks is severely hampered by the vanishing gradient problem \citep{hochreiter1997long}.

Furthermore, the difficulty of modelling the parity function drastically escalates with bitstring length \citep{hahn2020theoretical}. Since flipping a single bit alters the parity, processing long bitstrings requires persistent and perfect memorisation. RNNs and LSTMs are argued to be fundamentally ill-suited for this, as tiny memory errors can lead to disastrous consequences \citep{zhao2020rnn}. While parity is naturally permutation invariant, and RNNs can be regularised towards this property to improve performance \citep{cohen2020regularizing}, a fundamental representational bottleneck remains.

Modern architectures are similarly challenged. Self-Attention networks (Transformers \citep{vaswani2017attention}) have been shown to have strong theoretical limitations in evaluating logical formulas (like parity). Asymptotic analysis suggests that they will fail to model the parity task correctly when the input sequence is sufficiently long \citep{hahn2020theoretical}.

While the adaptive computation approach of PonderNet \citep{banino2021pondernet} achieved near-perfect accuracy, this success highlights the increasing resource demand. It suggests that as inputs become more complex, the required computational resources become enormous \citep{abbe2023provable}, and crucially, the model lacks the ability to extrapolate to input lengths beyond those seen during training.

\section{Revisiting AlphaZero and LCZero}
\label{sec:impartial_revisiting}

In this section, we illustrate that a distinct set of board positions, particularly those whose optimal move selection hinges on employing a parity function or similar logical structures, poses significant challenges for AlphaZero-style algorithms. We demonstrate this by examining specific chess positions where Leela Chess Zero (LCZero) exhibits difficulties in identifying the winning move due to parity-related intricacies. This analysis provides crucial context for understanding the more pronounced limitations observed in purely impartial games like Nim.

AlphaZero, while groundbreaking, was not released as open-source. However, several community-driven self-play RL projects, such as LCZero for chess, Leela Zero and KataGo \citep{wu2019accelerating} for Go, and AobaZero\footnote{\texttt{AobaZero}: \url{http://www.yss-aya.com/aobazero/index_e.html}} for shogi, have successfully replicated and extended the core principles of AlphaZero \citep{cazenave2020polygames}. These projects leverage distributed computing, relying on contributions from a global community of enthusiasts, enabling the training of remarkably powerful game-playing agents. Notable open-source variants also include the Elf Go project \citep{tian2019elf} and a publicly released variant of AlphaZero \citep{wu2019accelerating}.

AlphaZero and LCZero often approach position evaluation and move generation in ways that fundamentally differ from human grandmasters or traditional, hand-coded chess engines. Their revolutionary evaluations and surprising moves have profoundly impacted the chess community \citep{sadler2019game}. This section focuses on a highly trained version of LCZero\footnote{Version: Last T60: 611246 (384 $\times$ 30)}, which utilises a neural network architecture with 30 residual blocks and 384 filters, surpassing AlphaZero's initial $20 \times 256$ configuration. LCZero additionally incorporates Squeeze-and-Excitation layers into its residual blocks and supports endgame tablebases \citep{maharaj2021chess}, potentially enabling it to exceed the original AlphaZero's capabilities. Earlier LCZero versions (e.g., $20 \times 256$ architecture) are generally weaker than these more advanced iterations.

For comparative analysis, we also utilise open-source Stockfish 14, an evolution of the traditional Stockfish engine (initially developed by Tord Romstad, Marco Costalba, and Joona Kiiski). Unlike its predecessor, Stockfish 14 integrates an Efficiently Updatable Neural Network (NNUE) for position evaluation \citep{nasu2018efficiently}, blending traditional search with neural network capabilities. While game-playing strength is commonly measured by Elo rating (detailed in Section \ref{subsec:impartial_elo}), alternative metrics assessing move quality rather than just game outcomes have also been proposed \citep{regan2011intrinsic}. As of August 7, 2021, Stockfish 14 held an Elo rating of 3555, Stockfish 8 3375, and LCZero 3336 on popular engine rating lists. LCZero, when run on comparable hardware to AlphaZero against Stockfish 8 with a fixed time control (1 minute per move), replicated AlphaZero's impressive performance.

Although the original AlphaZero source code remains proprietary, detailed evaluations of its moves on select chess positions have been publicly documented \citep{sadler2019game}. We selected one such board position (Figure \ref{fig:LC0_1}) to directly compare AlphaZero's and LCZero's evaluations. In this position, the optimal move is \pawn d5. The policy network of our highly trained LCZero (using net T60: 611246 (384x30)) correctly prioritises \pawn d5, assigning it the highest prior probability (23.13\%), as shown in Figure \ref{fig:LC0_2}. AlphaZero, with a different training history, showed a preference for \bishop d3 but still identified \pawn d5 as a promising move. Figure \ref{fig:LC0_3} illustrates that after a modest number of MCTS simulations (visiting 64 or 256 nodes), both AlphaZero and LCZero converge on \pawn d5 as the best move, albeit with differing levels of confidence. Furthermore, as depicted in Figure \ref{fig:LC0_4}, the value networks of both agents consistently indicate a favourable position for White, with confidence increasing with more MCTS simulations.

\begin{figure}[htbp]
    \centering
    \setlength{\tabcolsep}{3pt}
    \subfloat[A chess board position where the white player is about to make a move.]{
        \label{fig:LC0_1}
        \newgame
        \fenboard{r1b2qk1/1pp3rp/4pnp1/1PP5/p2PBp2/P7/1BQ2P2/K1R1R3 w - - 0 1}
        \scalebox{0.7}{\showboard}
    }
    \hspace{1em}
    \subfloat[Prior probabilities from the policy network of AlphaZero and LCZero for the top moves]{
       \label{fig:LC0_2}
        \begin{tabular}{lcccccccc}
        \toprule
        \textbf{Move}  &  \bishop d3 & \bishop f3 & \pawn c6 & \pawn d5 & \bishop g2 & \pawn f3 & \bishop h1 & \queen c4\\
        \midrule
        \textbf{Prior prob (AlphaZero)} & \textbf{29.77}\% & 18.82\% & 16.15\% & \textbf{10.21}\% & 4.75\% & 3.5\% & 4.75\% & 1.2\% \\
        \midrule
        \textbf{Prior prob (LCZero)} & 6.01\% & 12.36\% & 16.27\% & \textbf{23.13}\% & 1.74\%  & 3.73\% & 1.41\% & 8.68\%\\
        \bottomrule
        \end{tabular}
    }
    \hspace{1em}
    \subfloat[The win probabilities from the value networks of AlphaZero and LCZero after visiting 64 and 256 nodes (that is equivalent to running 64 and 256 MCTS simulations.)]{
    \label{fig:LC0_3}
    \begin{tabular}{lccccccc}
        \toprule
        \textbf{Move}  &  \bishop d3 & \bishop f3 & \pawn c6 & \pawn d5 & \bishop g2 & \pawn f3 & \bishop h1 \\
        \midrule
        \textbf{Win prob (AlphaZero 64 nodes)} & 60.1\% & 64.5\% & 77.3\% & \textbf{87.1}\% & 61.6\% & 67.3\% & 61.6\%  \\
        \midrule
        \textbf{Win prob (AlphaZero 256 nodes)} & 60.1\% & 64.5\% & 77.7\% & \textbf{83.1}\% & 61.6\% & 67.3\% & 61.6\%  \\
        
       \midrule
        \textbf{Win prob (LCZero 64 nodes)} & 62.8\% & 62.8\% & 71.2\% & \textbf{71.6}\% & 55.7\% & 55.7\% &55.7\% \\
        \midrule
        \textbf{Win prob (LCZero 256 nodes)} & 59.0\% & 62.2\% & 63.3\% & \textbf{67.8}\% & 50.0\% & 58.2\% & 50.0\%  \\
        \bottomrule
    \end{tabular}
        }
        \hspace{1em}
    \subfloat[Comparison after running a given number of MCTS simulations. Across all evaluations, d5 consistently emerged as the optimal move.]{
        \label{fig:LC0_4}
        \begin{tabular}{lccccc}
            \toprule
            \textbf{Number of nodes visited} &  64 & 256 & 1024 & 65536 & 4194304  \\
            \midrule
            \textbf{AlphaZero eval} & 65.7\% & 74.1\% & 71.6\% & 67.9\% & \textbf{73.5}\%  \\
            \midrule
            \textbf{LCZero eval} & 70.1\% & 65.0\%  & 67.4\% & 68.3\% & \textbf{73.1}\% \\
            \bottomrule
        \end{tabular}
        }
    
    \caption{\label{fig:LC0} LCZero and AlphaZero evaluations on a standard chess position. For the chess board position in (a), both LCZero and AlphaZero promptly agree that \pawn d5 is the optimal move, despite slight differences in their initial PV-NN evaluations (b). After MCTS simulations, both agents consistently identify \pawn d5 as the best line (c) and show converging win probabilities for White (d).}
\end{figure}

As shown in Figure \ref{fig:LC0_4}, with an increasing number of MCTS simulations, AlphaZero's win probability initially showed a slight dip as it explored various defensive lines, before climbing again. Ultimately, after examining more than four million nodes, both engines yielded nearly identical evaluations, with win probabilities of 73.5\% and 73.1\%, respectively. This demonstrates that LCZero effectively replicates AlphaZero's capabilities in chess \citep{sadler2019game}. Even moves selected by LCZero without extensive search (i.e., greedily based on PV-NN evaluations) perform remarkably well, occasionally defeating strong human players.

However, a critical distinction arises when comparing RL agents with human intuition. While humans typically consider far fewer moves, human pattern recognition can sometimes identify features that LCZero's PV-NNs might overlook. Consider the position in Figure \ref{fig:LC1}. Any competent chess player would intuitively recognise that White has a forced mate in two moves (1.\queen g8+, \rook g8; 2.\knight f7+ mate) without extensive calculation. LCZero's policy network, however, initially prioritises \knight f7, leading to a detour in its search. While LCZero quickly finds the winning sequence (in less than a millisecond), this example highlights that its PV-NN, in isolation, does not always evaluate all positions with perfect accuracy. This imperfection becomes critically relevant for impartial games, where even minor perturbations can completely disrupt the correlation between PV-NN evaluations and correct game outcomes. This aligns with findings by \cite{lan2022alphazero}, who noted that KataGo lacks general knowledge applicable to all legal states and is susceptible to adversarial perturbations that, despite not altering the optimal move, can mislead the agent into suboptimal play.

\begin{figure}[htbp]
    \centering
    \subfloat[A chess board position]{
        \newgame
        \fenboard{1rr4k/6pp/7N/3Q4/8/qBp5/P1P5/1K6 w - - 0 1}
        \scalebox{0.65}{\showboard}
    }
    \subfloat[LCZero evaluations for the top 4 moves]{
        \begin{tabular}{lccc}
        \toprule
        \textbf{Move}  &  \knight f7 & \queen g8 & \queen d8  \\
        \midrule
        \textbf{Prior Prob} & 60.69\% & 15.96\% & 2.50\%  \\
        \midrule
         \textbf{V-value} & -0.276 & 0.588 & -0.968 \\
        \midrule
        \textbf{Win prob} & 36.2\% & 79.4\% & 1.6\% \\
        \midrule
         \textbf{Q-value (15 nodes)} & -0.075\% & 1.0\% & 0.0\% \\
        \midrule
        \textbf{Win prob (15 nodes)} & 48.5\% & 100\% & 0\% \\
        \bottomrule
        \end{tabular}
    }
    \caption{\label{fig:LC1} LCZero's initial evaluation in a forced-winning chess position. In this chess position, White has a forced win in two moves, involving a queen sacrifice leading to checkmate. LCZero's initial PV-NN evaluation (without search) does not immediately identify the forced win. Only after 11 MCTS nodes does the search shift focus from \knight f7 to the winning move \queen g8, and the forced mate is found after 15 nodes.}
\end{figure}

The next example in Figure \ref{fig:LC2} further illustrates the difficulty AlphaZero-style algorithms face with parity-related problems. This chess position is constructed to mimic a Nim position: the right side of the board is a losing position for any player who moves, while the left side is equivalent to a Nim game with two heaps of three and two counters respectively, where only removing one or two counters from a heap is allowed (sometimes referred to as "Bogus Nim"). The optimal move in this Nim variant is to remove one counter from the heap of three. Analogously, the winning move for White in the chess position is \pawn c3-c4. However, LCZero's policy network initially suggests \pawn a2-a4 with a 50.4\% prior probability, a move that disastrously leads to a losing position. While the value network correctly identifies \pawn c4 as favourable (90.3\% win probability), the policy network's initial strong preference for a sub-optimal move indicates a challenge in accurately prioritising parity-dependent moves. Despite this initial misdirection, LCZero's MCTS quickly corrects its course, finding the optimal sequence after only a few nodes.

\begin{figure}[htbp]
    \centering
    \subfloat[A chess board position]{
        \newgame
        \fenboard{6k1/2p5/5P1P/p7/8/2P2p1p/P7/6K1 w - - 0 1}
        \scalebox{0.65}{\showboard}
    }
    \subfloat[LCZero's evaluations for PV-NN for the top 4 moves]{
        \begin{tabular}{lcccc}
        \toprule
        \textbf{Move}  & \pawn a4 & \pawn c4 & \pawn a3 & \pawn f7 \\
        \midrule
        \textbf{Prior prob} & \textbf{50.4}\% & \textbf{20.7}\% & 15.4\% & 4.4\%  \\
        \midrule
        \textbf{V-value} & -0.632 & 0.806 & -0.817 & -0.88\% \\
        \midrule
        \textbf{Win prob} & 18.4\% & 90.3\% & 9.2\% & 6.0\%  \\
        \midrule
         \textbf{Q-value (3 nodes)} & -0.632 & \textbf{0.806} & -0.248 &  -0.248 \\
        \midrule
        \textbf{Win prob (3 nodes)} & 18.4\% & \textbf{90.3}\% & 37.6\%  & 37.6\% \\
        
        \bottomrule
        \end{tabular}
    }
    
    \caption{\label{fig:LC2} LCZero's struggles with parity-dependent chess positions. This chess position is designed to mimic a Nim variant (Bogus Nim on $[3,2]$). While White's winning move is \pawn c3-c4, LCZero's policy network gives a sub-optimal move (\pawn a2-a4) the highest prior probability (50.4\%). Despite the value network correctly identifying the optimal move's potential, this highlights the PV-NN's initial misguidance on parity-related structures. MCTS quickly finds the correct move after a few nodes.}
\end{figure}

Despite being equipped with highly sophisticated PV-NNs, LCZero generally struggles to accurately evaluate positions with strong parity-related properties, including scenarios involving "zugzwang," "waiting moves," and "triangle manoeuvres," which are common themes in chess endgames. In standard gameplay, these issues are often mitigated because the extensive MCTS compensates for initial PV-NN inaccuracies.

However, some examples demonstrate that the policy network can occasionally be 'blind' to crucial, non-obvious moves that MCTS may fail to discover even after extensive search. Consider the chess position in Figure \ref{fig:LC3}, which arose in a game between Stockfish 12 and LCZero. White has a forced checkmate in five moves, starting with \rook c2. LCZero's policy network, however, assigned the highest prior probability to \rook a5+. A more critical problem is that LCZero's policy network failed to effectively guide the MCTS, resulting in the agent being unable to find the winning sequence even after exploring over one million nodes. This failure underscores a limitation inherent to using MCTS, as traditional alpha-beta search, commonly employed in conventional chess engines, would typically find such a forced mate almost instantly. This reinforces the concept of "blind spots" \citep{lan2022alphazero, wu2019accelerating}, where critical information is entirely missed, indicating a lack of robust generalisation across all legal states. These observed weaknesses in chess, particularly concerning parity and complete oversight of optimal lines, serve as a strong motivation for our investigation into impartial games, where such structures are even more fundamental to game logic.

\begin{figure}[htbp]
    \centering
    \subfloat[A chess board position]{
        \newgame
        \fenboard{6r1/5p2/1r6/2kpPb2/Rp1p4/3P2P1/1R3PK1/3B4 w - - 0 1}
        \scalebox{0.65}{\showboard}
    }
    \hspace{0.1em}
    \subfloat[LCZero evaluation's from PV-NN for the top 4 moves]{
        \begin{tabular}{lcccc}\hline
            \toprule
            \textbf{Move} & \rook a5 & \rook c2 & \bishop e2 &  \rook b3 \\
            \midrule
            \textbf{Prior prob} & \textbf{35.9}\% & 16.5\% & 10.3\% & 4.7\%  \\
            \midrule
            \textbf{V-value} & 0.29 & 0.27 & -0.005 & 0.06  \\
            \midrule
        \textbf{Win prob} & 64.5\% & 63.5\% & 49.8\% & 53.0\%  \\
            \bottomrule
        \end{tabular}
        \vspace{3em}
    }
    \caption{\label{fig:LC3} LCZero's failure to find a forced checkmate. LCZero blundered in this chess position, failing to recognise a forced checkmate in five moves starting with \rook c2. The policy network prioritised the sub-optimal \rook a5+ move (a). Even after over one million MCTS nodes, LCZero could not discover the winning sequence, highlighting a 'blind spot' where critical moves are entirely missed, a significant limitation compared to traditional search algorithms.}
\end{figure}

\section{Two levels of mastery}
\label{sec:impartial_two}

In the domain of combinatorial game theory, a two-player game is considered \textit{solved} if its optimal outcome can be predicted from any position, assuming both players make optimal moves \citep{allis1994searching}. Game solutions are typically categorised into three levels. An \textit{ultra-weak solution} determines only whether the first player has a winning advantage from the initial position \citep{van2002games}. Given that self-play RL agents need to generalise and make effective moves from arbitrary positions encountered during gameplay, the ultra-weak solution concept is less relevant to our investigation. Instead, we focus on the more comprehensive definitions of game solutions, as summarised in Table \ref{tab:level_solved}.

\begin{table}[h]
\caption{Traditional Levels of Solving a Two-Player Game}
\label{tab:level_solved}
\centering
\begin{tabular}{l p{.70\linewidth}}
\toprule
\multicolumn{1}{c}{Level} & \multicolumn{1}{c}{Description} \\ [0.5ex]
\midrule
\textbf{Weakly solved} & The algorithm has learned to make a sequence of good moves that, starting from the initial position, consistently lead to a win against any opponent in actual gameplay. \\
\textbf{Strongly solved} & The algorithm has learned to make optimal moves in \textit{all possible game positions} that can arise through legal play from the initial position. \\
\bottomrule
\end{tabular}
\end{table}

Building upon these established definitions, we propose two distinct measures to assess the extent of mastery achieved by an RL algorithm: \textit{champion} and \textit{expert}. These measures characterise the depth and breadth of an agent's understanding and strategic capability within a game, as delineated in Table \ref{tab:level_mastery}.

\begin{table}[h]
\caption{Our Proposed Measures of RL Agent Mastery}
\label{tab:level_mastery}
\centering
\begin{tabular}{l p{.70\linewidth}}
\toprule
\multicolumn{1}{c}{Measure} & \multicolumn{1}{c}{Description} \\ [0.5ex]
\midrule
\textbf{Champion} & A player's ability to achieve optimal (winning or drawing) results against any opponent when the game is initiated from its standard initial position, often by steering the game into familiar sub-spaces. \\
\textbf{Expert} & A player's ability to consistently execute the optimal move in \textit{any legal position} that can arise from play, demonstrating a comprehensive and robust understanding of the game's entire state space. \\
\bottomrule
\end{tabular}
\end{table}

This distinction is of paramount importance for evaluating the true generalisation capabilities of RL agents. The question posed by \cite{lan2022alphazero}, regarding whether an RL agent truly learns "general knowledge applicable to any legal game state," directly pertains to its capacity to become an \textit{expert}. A champion agent might possess a winning strategy from the initial state by effectively guiding the game into its "comfort zone," where its knowledge is deep and reliable. In contrast, an expert agent must consistently make optimal decisions across the entire relevant game graph, irrespective of whether the position is common or rare. While it is beyond the scope of this paper to provide a formal proof, it is intuitively plausible that in some games, becoming a champion without simultaneously becoming an expert is impossible, especially when the winning player cannot sufficiently constrain the opponent's options to force the game into well-understood territories. A more rigorous theoretical analysis of the conditions under which champion-level play necessitates expert-level knowledge is an intriguing avenue for future research.

The concept of "problem chess," involving artificially composed chess positions designed to test specific tactical or strategic motifs, offers a relevant analogy. LCZero, like AlphaZero, is not explicitly trained to solve such artificial problems, and it is therefore unsurprising that it can struggle with them \citep{maharaj2021chess}. The primary objective of AlphaZero-style algorithms in chess, shogi, and Go has been to learn to play and win games from standard starting positions---a "champion" notion of mastery. Their success is measured by their ability to achieve superior game outcomes rather than by their capacity to find optimal moves in every conceivable (and potentially rare or adversarial) situation, which aligns with our "expert" definition. Our work explores the extent to which self-play RL, trained for champion-level performance, can truly generalise to expert-level play, particularly in games with a strong parity structure like Nim.

The champion and expert concepts, and their implications for RL agent performance, will be thoroughly discussed in conjunction with our experimental results in the next section. To further clarify this critical distinction, we offer the following analogy and a specially crafted Nim board example.

\medskip
\noindent
\textbf{Example 1: The Tactical Fighter Analogy.} Imagine a combat agent capable of forcing its opponent to engage in one of two distinct environments: a wide-open savanna or a dense jungle. A champion-level fighter might be an absolute expert in savanna combat but utterly inept in the jungle. Nevertheless, this champion can consistently win by always compelling the fight into the savanna. To be deemed an "expert" fighter, the agent would need to master both savanna and jungle combat, ensuring optimal performance in any situation it might encounter, regardless of the environment. This illustrates how an agent can achieve optimal outcomes from initial conditions (champion) without possessing comprehensive knowledge of all possible sub-problems (expert).

\medskip
\noindent
Here is another simple, yet extreme, example highlighting the relevance of these two notions from the game of Nim.

\medskip
\noindent
\textbf{Example 2: The $[2, 1, \ldots, 1]$ Nim Board.} Let $N \in \mathbb{N}$ and consider a Nim board configured as $[2, 1, 1, \ldots, 1]$, where there is one heap of 2 counters and $N$ heaps of 1 counter, with $N$ being a large number (e.g., $N=100$). An RL agent is likely to become a champion in this specific game configuration after relatively little training. It can quickly learn the optimal opening moves: specifically, to remove either 1 or 2 counters from the initial heap of 2, thus bringing the nim-sum to zero and achieving a P-position. Through self-play, the agent effectively memorises these few crucial initial moves.

However, the task of selecting the best move in a general position derived from this board, represented as $[v_1, v_2, \ldots, v_{N+1}]$ where $v_1 \in \{0,1,2\}$ and $v_j \in \{0,1\}$ for $j \in \{2, \ldots, N+1\}$, is fundamentally equivalent to evaluating the parity of the sub-configuration $[v_2, \ldots, v_{N+1}]$. This involves computing the XOR sum of $N$ binary values. Due to the inherent challenges of neural networks in modelling parity functions, especially without a strong incentive to learn such a generalizable representation during self-play, a champion agent for this specific Nim board is not expected to simultaneously become an expert agent across all possible derivative states. This example underscores how an agent can achieve winning performance in specific initial conditions without developing a general, expert-level understanding of the underlying mathematical properties that govern all possible game states.

\section{Reinforcement learning for Nim}
\label{sec:impartial_rl}

The AlphaZero-style learning paradigm has achieved unprecedented success in complex board games, with various open-source implementations available\footnote{\texttt{AlphaZero-general}: \url{https://github.com/suragnair/alpha-zero-general}; \texttt{AlphaZero\_Gomoku}: \url{https://github.com/junxiaosong/AlphaZero_Gomoku}}
and specialized RL libraries offering optimized frameworks \citep{lanctot2019openspiel, liang2018rllib}. However, for our specific diagnostic analysis of AlphaZero-style algorithms on impartial games, these existing solutions lacked critical functionalities. Our objectives necessitated granular control over the training process and detailed insights into the network's internal representations, including, but not limited to, the real-time calculation of agent Elo ratings against its ancestors and, crucially, the precise evaluation of the policy network's output accuracy against the known optimal moves derived from the nim-sum function as training progresses. These diagnostic capabilities are pivotal for understanding the learning dynamics in domains where optimal play is mathematically defined.

This necessitated the implementation of a custom AlphaZero-style self-play algorithm. Our implementation leverages PyTorch \citep{paszke2019pytorch} for neural network training and Ray \citep{moritz2018ray} for efficient parallelisation of MCTS simulations. We made considerable efforts to ensure that our implementation rigorously replicates the architectural and algorithmic details of the original AlphaZero algorithms as described in \cite{silver2017mastering} and \cite{silver2018general}. While maintaining this high fidelity, necessary adaptations to the neural network architectures were made to accommodate the specific input and output representations of Nim, which will be elaborated in detail in Section \ref{sec:impartial_implementing}.

\subsection{Implementing an AlphaZero-style algorithm for Nim}
\label{sec:impartial_implementing}

The AlphaZero algorithm initiates training from a blank slate state, with its sole prior knowledge being the game's fundamental rules \citep{silver2018general}. Its architecture relies on three interdependent components: the policy network, the value network, and the MCTS. The policy network, typically a deep neural network, outputs a probability distribution $P(s, a)$ over all legal actions $a \in \mathcal{A}$ for a given state $s$. Probabilities for illegal actions are set to zero, and the remaining probabilities are re-normalised. This network serves to narrow the search space by prioritising actions with a high likelihood of leading to favourable outcomes, thereby reducing the MCTS's breadth. Concurrently, the value network outputs a scalar value $v \in [-1, 1]$, providing an estimate of the expected outcome from state $s$. A higher value indicates a greater probability of the current player winning, and vice versa. By predicting the value of leaf nodes, the value network effectively prunes the search tree's depth, as a full rollout to the game's conclusion is computationally expensive and its true outcome (1 for win, -1 for loss) is only definitively known at the end of the game.

The training targets for the value network are derived from the actual outcomes of self-play games. If the agent wins a game, all encountered positions within that game are labelled as winning states (target value +1), and conversely, losing positions are labelled as -1. This self-generated labelling process is susceptible to noise, especially during the early stages of training when the PV-NN is untrained or weakly trained. Such noisy labels pose a significant challenge for neural networks, particularly in learning abstract functions like parity on long bitstrings \citep{zhou2023exploring}. Consequently, this inherent difficulty extends directly to the PV-NN's ability to accurately evaluate positions in Nim, where the optimal strategy is fundamentally linked to the nim-sum.

Recognising that Long Short-Term Memory (LSTM) networks are theoretically capable of modelling perfect parity functions due to their ability to learn long-range dependencies in sequential data \citep{cohen2020regularizing}, we employed an LSTM-based architecture for our PV-NN. The policy and value networks share a common LSTM layer, which then branches into two separate heads: a policy head and a value head. For Nim games with 5, 6, and 7 heaps, the shared LSTM layer used a hidden size of 128. While it is intuitive to scale network depth with increasing board size, we empirically found that employing a larger number of LSTM layers was detrimental to performance, often destabilising the training process. The output of the shared LSTM layer feeds into the two heads. The policy head consists of 25, 36, and 49 nodes for 5, 6, and 7 heaps Nim, respectively (matching the action space), with its logits passed through a softmax function to yield a probability distribution over actions. The value head comprises a single node, outputting a scalar value passed through a tanh activation function to ensure its range is confined to $[-1, 1]$.

The MCTS commences with a root node representing the current game state. Each node in the tree corresponds to a state, and each edge represents an action. The tree is built incrementally through a predefined number of simulations, each starting from the root and traversing to a leaf node by selecting actions according to the Upper Confidence Bound for Trees (UCT) formula. Specifically, actions are selected using the following equation:
\begin{equation}
    a_t = \argmax_a(Q(s,a)+U(s,a))
    \label{eqn:action_selection}
\end{equation}
where $Q(s,a)$ is the mean action value, calculated as:
\begin{equation}
\label{eqn:q_value}
    Q(s, a) = \frac{1}{N(s, a)}\sum_{s'|s, a\rightarrow s'}V(s')
\end{equation}
Here, $N(s, a)$ denotes the visit count for action $a$ from state $s$, and $V(s')$ is the value of the end state $s'$ of a simulation (either from the value network if $s'$ is an intermediate state, or the game outcome if terminal). The exploration term $U(s,a)$ plays a critical role and is calculated using:
\begin{equation}
\label{eqn:puct}
    U(s,a)=c_{put}P(s,a)\dfrac{\sqrt{\sum_b{N(s,b)}}}{1 + N(s,a)}
\end{equation}
The constant $c_{put}$ controls the balance between exploitation and exploration. While AlphaGo \citep{silver2016mastering}, AlphaGo Zero \citep{silver2017mastering}, and AlphaZero \citep{silver2018general} left $c_{put}$ unspecified or tuned it empirically for specific games, we found its value significantly impacted performance on Nim. Setting $c_{put}$ too low excessively discourages exploration, while setting it too high diminishes the action value's influence, impairing search depth effectiveness. Initial attempts with values common in other domains, such as $c_{put}=\{1, 1.5, 2, 3\}$ (where $c_{put}=1.5$ yielded satisfactory results for Elf OpenGo \citep{tian2019elf}), performed poorly for Nim. Consequently, we adopted a dynamically adjusted exploration term inspired by MuZero \citep{schrittwieser2020mastering}:
\begin{equation}
\label{eqn:u_s_a}
    U(s,a)=P(s,a)\cdot\dfrac{\sqrt{\sum_b{N(s,b)}}}{1 + N(s,a)}\left(c_1 + \log\left(\dfrac{\sum_b{N(s,b)} + c_2 + 1}{c_2}\right)\right)
\end{equation}
where $c_1=0.25$ and $c_2=19652$. This formulation provides a more sophisticated balance.

To further encourage broad exploration from the root node, Dirichlet noise was added to the prior probabilities $P(s,a)$. This noise is indispensable for ensuring the search tree is sufficiently branched, thereby preventing the MCTS from becoming overly fixated on moves with initially high prior probabilities, which might be incorrect.
\begin{equation}
   P(s,a) \leftarrow (1-\epsilon) \cdot P(s,a) + \epsilon \cdot \eta_{a}
    \label{eqn:add_noise}
\end{equation}
Here, $\eta_a$ is sampled from a Dirichlet distribution Dir($\alpha$), with $\alpha$ set to 0.35. The constant $\epsilon$ is set to 0.25 during training but to 0 during evaluation to eliminate the noise effect. The typical $\alpha$ values used in AlphaZero for Chess, Shogi, and Go are 0.3, 0.15, and 0.003, respectively. Given that the average number of legal moves in the Nim variants we study is generally lower than in Chess, we opted for a higher $\alpha$ value of 0.35 to promote more diverse exploration relative to the smaller action space. While theoretically $\alpha=0.5$ might be optimal for maximising entropy, our empirical fine-tuning found $\alpha=0.35$ to yield better practical outcomes.

The MCTS selection policy, as evident from Equation \ref{eqn:action_selection}, jointly considers the action value $Q(s,a)$, the visit count $N(s,a)$, and the prior probability $P(s,a)$ from the policy network. Actions with lower visit counts, higher prior probabilities, and higher estimated values are preferentially chosen. Consequently, when the policy network fails to assign a high prior probability to optimal moves, the search is fundamentally misguided towards suboptimal nodes, severely impairing its effectiveness. Analogously, an inaccurate value network can lead to premature truncation of promising search paths, preventing the discovery of winning states.

After completing a predefined number of simulations for each move, the MCTS returns a probability distribution $\boldsymbol\pi(\textbf{a}|s)$ over all actions, from which an action is sampled for execution. This distribution is calculated as:
\begin{equation}
    \boldsymbol\pi(\textbf{a}|s) = \dfrac{N(s,\textbf{a})^{1/\tau}}{\sum_{b}N(s,b)^{1/\tau}}
    \label{eqn:pi_action_selection}
\end{equation}
Here, $\tau$ represents the temperature parameter, which modulates the exploration/exploitation balance. We set $\tau=1$ for the first 3 moves of a game, allowing action selection to be proportional to their visit counts, thus encouraging broader exploration in the opening. For subsequent moves, $\tau=0$ is used, ensuring that the action with the highest visit count is deterministically chosen (pure exploitation). During the evaluation phase, $\tau$ is consistently set to 0 for all moves. These posterior probabilities $\boldsymbol\pi(\textbf{a}|s)$ serve as the training targets for the policy network, aiming to refine its initial predictions.

It is worth noting that while our study employs a single-frame state representation for Nim, it is crucial to note the distinction from games where history is integral to the state. Unlike Chess, Go, or Shogi, where multi-frame representations are necessary to handle repetitions, cycles, and complex rule enforcement, Nim is inherently memoryless. Its game state is fully determined by the current heap configuration, rendering past moves irrelevant. This property simplifies state representation for Reinforcement Learning. For the scope of this work, focusing on the fundamental challenges Nim's combinatorial nature poses to AlphaZero, a single-frame representation is both sufficient and advantageous, eliminating the complexities of history-dependent state constructions.

\subsection{Experiment setup and results}
\label{sec:impartial_experiments}

This section details the experimental configurations and presents the results derived from training our AlphaZero-style algorithm on various Nim games. The performance of the trained agents is subsequently analysed in depth, drawing upon the 'champion' and 'expert' perspectives introduced earlier.

Our implementation of the PV-NN employs a shared LSTM layer with separate policy and value heads. This architecture was applied across all Nim variants (5, 6, and 7 heaps). While the number of nodes in the output heads was adapted to match the respective action spaces, the core network architecture remained uniform. We chose a high number of MCTS simulations for each move, increasing this number with growing heap count. This approach not only aims to yield more robust heuristics but also to mitigate the algorithm's known sensitivity to hyperparameter variations. During both training and evaluation, the number of simulations per move was set to $s=\{50, 60, 100\}$ for Nim games with $h=\{5, 6, 7\}$ heaps, respectively. All simulations were executed in parallel across 8 CPUs. The substantial increase in simulations for 7-heap Nim was a deliberate choice to rigorously ascertain whether insufficient search depth was a contributing factor to performance limitations, despite incurring significant computational costs.

The specific variant of Nim studied is described in Section \ref{sec:impartial_games}, where each heap initially contains an odd number of counters. For instance, the initial board for 5-heap Nim (as depicted conceptually in Fig. \ref{fig:nim_board}) consists of heap sizes $[1, 3, 5, 7, 9]$. The board representation provided as input to the neural network encodes the counters using unitary numbers (1 for present, 0 for removed), with -1 acting as a delimiter between distinct heaps. This provides a variable-length bitstring representation to the LSTM.

As shown in Table~\ref{tab:nim_spaces}, the state and action space sizes grow rapidly with the number of heaps. On average, the game length when moves are selected randomly by two players is approximately 10 for 5-heap Nim, 13 for 6-heap Nim, and 16 for 7-heap Nim. During training, we collected 100 episodes of self-play interaction data in the form of $(s, \pi, r)$ tuples (state, MCTS-derived policy target, game outcome reward) for updating the neural networks. The policy network was trained using cross-entropy loss, and the value network using mean squared error (MSE). The detailed performance analysis under the champion and expert perspectives, leveraging this experimental setup, is presented in the following sections.

\begin{table}[h]
\caption{State and action space sizes for different Nim board configurations used in our experiments.}
\centering
\begin{tabular}{@{}ccc@{}}
\toprule
\textbf{Number of Heaps} & \textbf{State Space Size} & \textbf{Action Space Size} \\
\midrule
5 & 3,840   & 25 \\
6 & 46,080  & 36 \\
7 & 645,120 & 49 \\
\bottomrule
\end{tabular}

\label{tab:nim_spaces}
\end{table}

\subsubsection{Champion measure by Elo rating}
\label{subsec:impartial_elo}
The Elo rating system, widely adopted in competitive two-player games such as chess, provides a robust methodology for ranking players based on their relative competitive strength. This constitutes a 'champion measure', reflecting an entity's ability to consistently defeat opponents. AlphaZero itself utilised Elo ratings to quantify its performance against formidable counterparts like AlphaGo Zero, AlphaGo Lee, and Stockfish. Given the absence of a pre-established competitive Nim agent with an associated Elo rating, we implemented a variation of the self-play Elo rating system \citep{tian2019elf}. This approach enables us to continuously monitor the relative strength of our agent and track its training progress, with the rating dynamically updated against all its historical predecessors to ensure an accurate reflection of its evolving competitiveness.

In our self-play Elo rating system, each newly trained agent is initialised with a baseline score of 1000. Following each training iteration, the newly trained agent is integrated into a repository of previously trained agents. Its Elo rating is then computed by conducting evaluation matches against all agents currently stored in this system. For any given match, let the agent undergoing training be denoted as Player A, and its opponent, one of its predecessors, as Player B. The expected score for Player A ($E_A$), representing its probability of winning the match, is calculated using the logistic function:
\begin{equation}
    E_A = \dfrac{1}{1 + 10^{(R_B-R_A)/400}}
\end{equation}
Similarly, the expected score for Player B ($E_B$) is:
\begin{equation}
    E_B = \dfrac{1}{1 + 10^{(R_A-R_B)/400}}
\end{equation}
For Nim, a drawing outcome is not possible; therefore, a match always concludes with either a win (score = 1) or a loss (score = 0) for Player A. If Player A wins the game, its Elo rating is updated according to the formula:
\begin{equation}
    R_A = R_A + K(1 - E_A)
\end{equation}
where $K$ represents the K-factor, which determines the maximum possible adjustment to a player's rating. Standard Elo practice typically employs a higher K-factor for new or less established players (leading to more volatile rating changes) and a lower K-factor for established, highly-rated players (for greater rating stability). Adhering to this principle, in our setting, the K-value for players who have participated in fewer than 20 evaluation games is set to 32, allowing for more rapid rating adjustments in early training phases. For players who have engaged in 20 or more games, indicating a more established and reliable rating, the K-value is reduced to 16 to ensure greater rating stability. Conversely, if Player A loses the game, Player B's Elo rating is updated as:
\begin{equation}
    R_B = R_B + K(0 - E_B)
\end{equation}

This self-contained approach negates any reliance on external programs or human expertise for ranking. However, a crucial implication of this methodology is that the derived Elo ratings are inherently relative to the pool of agents generated during self-play. Consequently, these ratings cannot be directly compared with Elo scores from external competitive systems or different agent populations. Nevertheless, these internal Elo ratings serve as a robust indicator of an agent's progressive improvement and competitive strength within the self-play ecosystem, establishing a valuable baseline for comparative analysis in future research on Nim AI.

\begin{figure}[ht]
\centering
\includegraphics[width=0.45\textwidth]{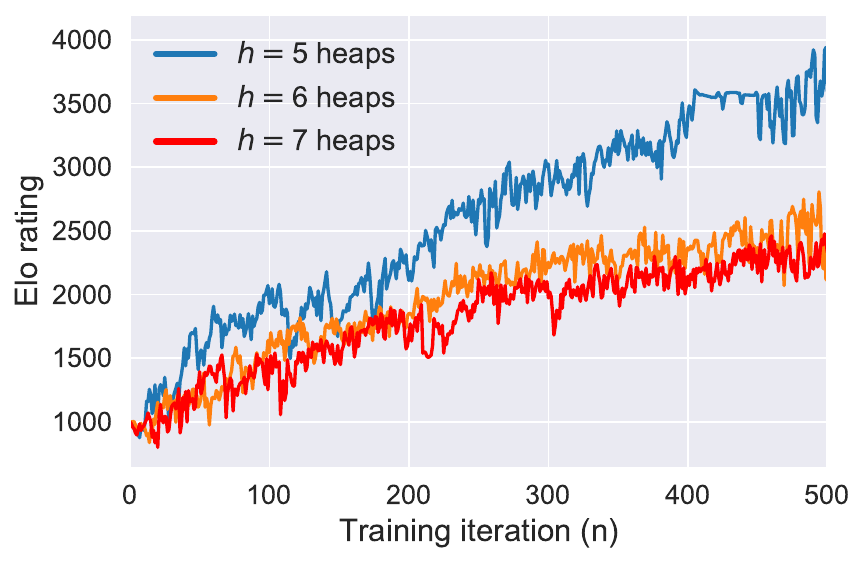}
\caption{\label{fig:elo_rating} The self-play Elo rating scores of the AlphaZero agent trained on Nim of 5, 6, and 7 heaps, respectively. Ratings were calculated at the end of every training epoch against all historical agents stored in the pool. The computational overhead for continuous Elo rating calculation during training is substantial; for the 7-heap Nim, the process required approximately 200 compute-hours with the specified configurations.}
\end{figure}

We meticulously monitored the self-play Elo rating of the AlphaZero agent across Nim variants with 5, 6, and 7 heaps. As illustrated in Figure \ref{fig:elo_rating}, the self-play Elo rating of the trained agent generally increases with training progress, signifying enhanced competitiveness. The AlphaZero agent for 5-heap Nim exhibited rapid Elo growth from the inception of training. In stark contrast, the growth for the 6-heap Nim agent was notably slower, and for the 7-heap Nim agent, the Elo rating became largely stagnant after approximately 420 iterations. This plateau indicates that while the agent may achieve some level of competitiveness, it encounters a significant performance ceiling that proves challenging to surmount. As will be demonstrated in the subsequent section, this critical bottleneck is directly attributable to the inherent inability of the PV-NNs to precisely and reliably evaluate Nim board positions, particularly as game complexity increases.

\subsubsection{Expert measure by the accuracy of PV-NN}
\label{subsec:impartial_performance}
Beyond assessing an agent's 'champion' capabilities through Elo ratings, a deeper understanding of its 'expert' mastery necessitates evaluating the intrinsic accuracy of its PV-NN. From an expert perspective, the policy network's action probability distribution should accurately reflect optimal moves. In Nim, the mathematically provable optimal moves are uniquely determined by the nim-sum property. Consequently, the policy network's accuracy is evaluated by comparing the probability assigned to the optimal move (as given by the nim-sum) against its actual prediction, specifically focusing on the most probable move. A similar policy evaluation metric, termed "Policy Top 1 Accuracy," has been employed in related works \citep{danihelka2022policy}.

\begin{figure}[ht]
\centering
\begin{subfigure}[b]{0.34\textwidth}
\centering
\includegraphics[width=\textwidth]{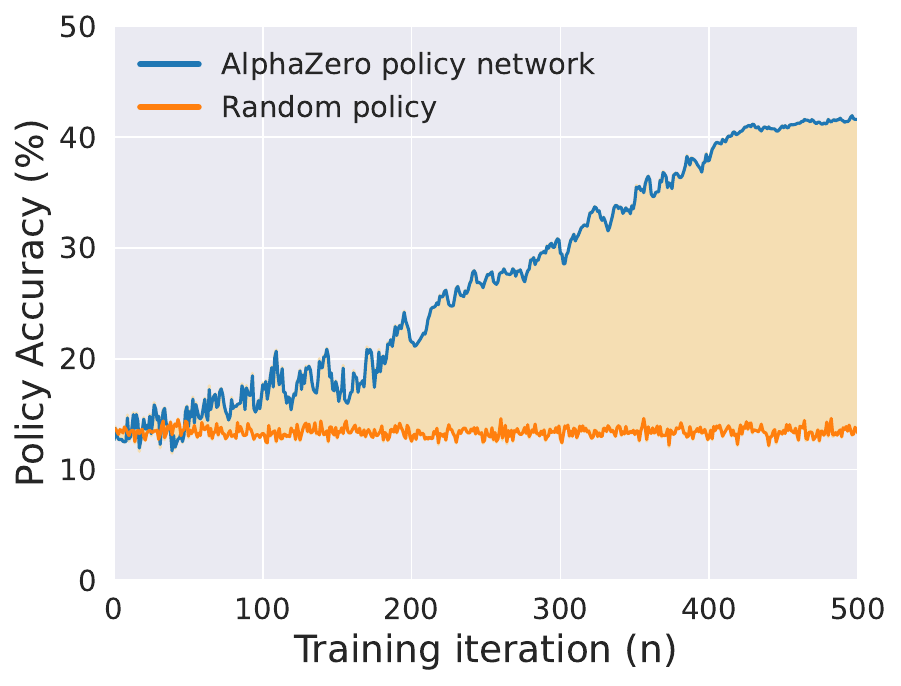}
\caption{$h=5$ heaps}
\end{subfigure}
\begin{subfigure}[b]{0.32\textwidth}
\centering
\includegraphics[width=\textwidth]{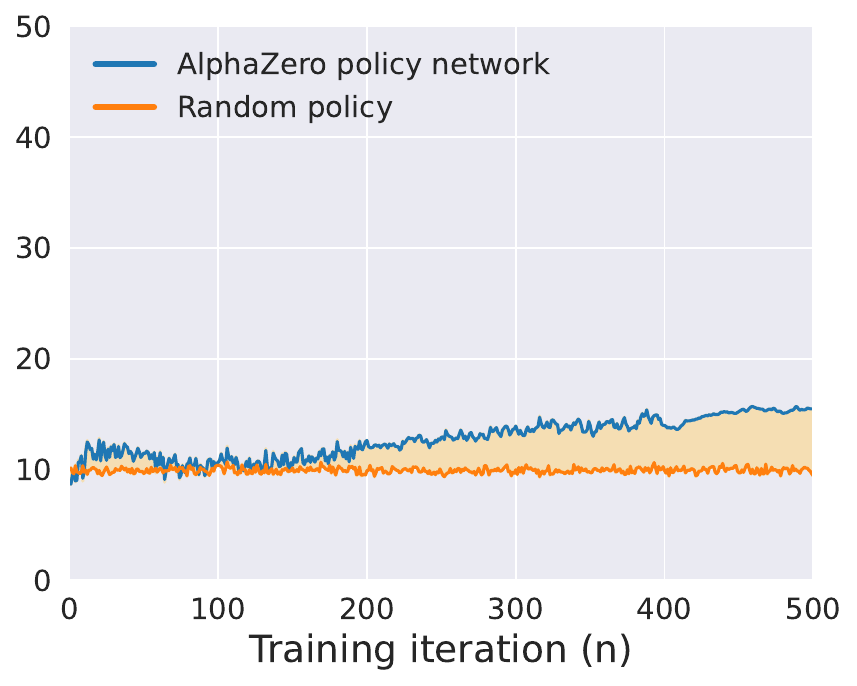}
\caption{$h=6$ heaps}
\end{subfigure}
\begin{subfigure}[b]{0.32\textwidth}
\centering
\includegraphics[width=\textwidth]{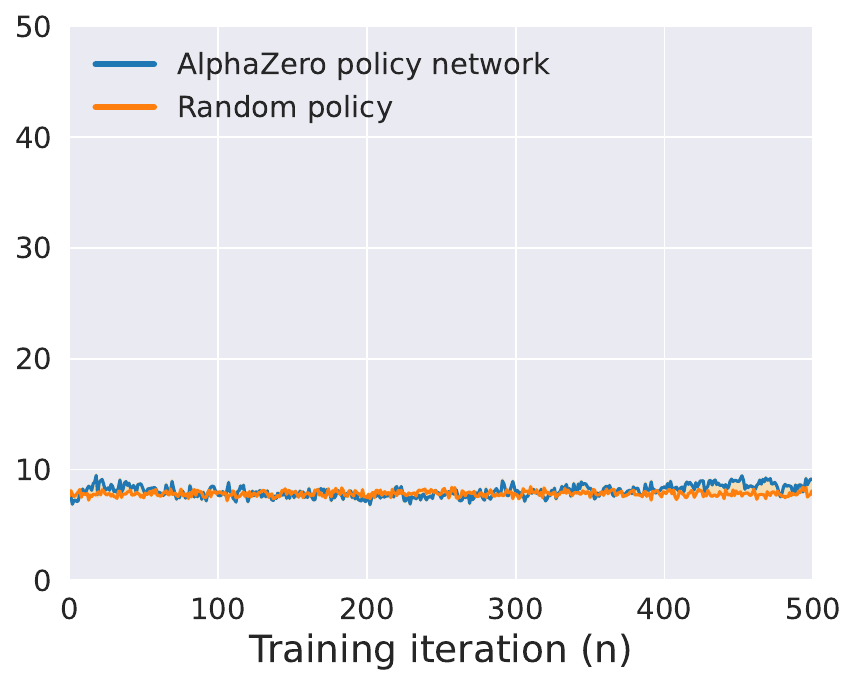}
\caption{$h=7$ heaps}
\end{subfigure}
\caption{\label{fig:alphazero_policy_accuracy} The accuracy of the policy network, measured against the ground-truth optimal moves, for Nim of 5, 6, and 7 heaps. The AlphaZero policy demonstrates superiority over the random policy on smaller boards, but its accuracy drastically drops as the board size grows, approaching random performance.}
\end{figure}

We established a random policy as a baseline for comparison with our AlphaZero policy. As depicted in Figure \ref{fig:alphazero_policy_accuracy}, the AlphaZero policy significantly outperforms the random baseline on 5-heap Nim. However, this advantage precipitously declines as the board size increases. For 7-heap Nim, the AlphaZero policy's Top 1 Accuracy becomes nearly indistinguishable from that of a random policy. This result is critical: while the policy network might still provide some minimal guidance to the MCTS that allows the agent to win some games (thus increasing its Elo), its fundamental inability to accurately identify optimal moves demonstrates a severe representational bottleneck. This highlights the neural network's struggle to implicitly learn the non-linear, parity-based optimal strategy from the raw state representation, particularly as the state and action spaces expand.

\begin{figure}[h]
\centering
\includegraphics[width=0.45\textwidth]{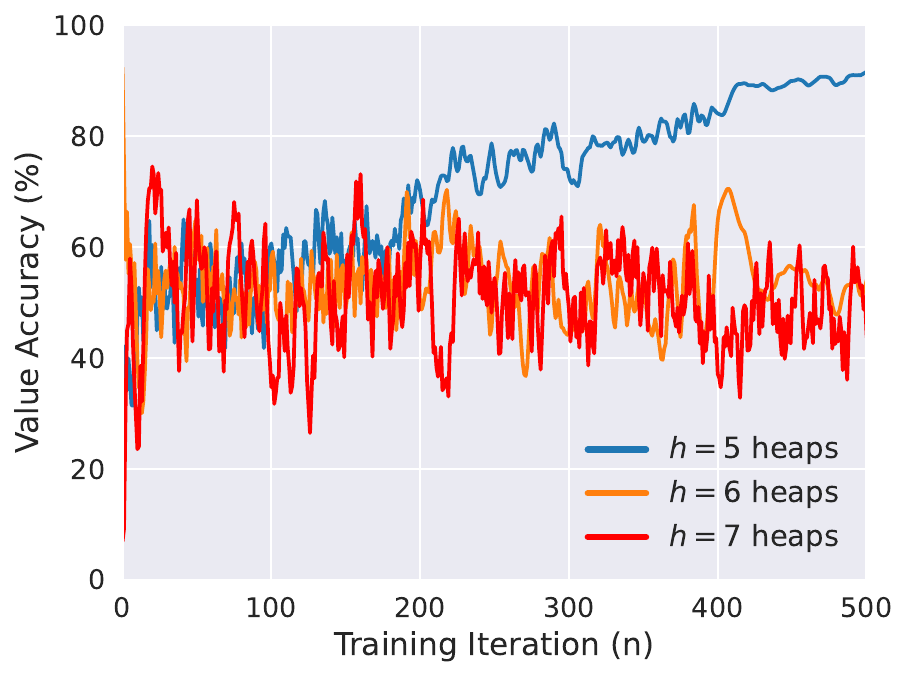}
\caption{\label{fig:alpha_value_accuracy} The accuracy of the value network in predicting winning/losing positions for Nim of 5, 6, and 7 heaps. Accuracy on 5-heap Nim consistently rises, reaching 90\% by 500 training iterations. For 6-heap Nim, accuracy exceeds 60\%, but for 7-heap Nim, it fluctuates near 50\%, indicating performance akin to random guessing.}
\end{figure}

The value network's role is to provide an accurate estimate of the game's expected outcome from a given state. In Nim, a winning position is one from which the current player can win, assuming optimal play. Conversely, a losing position is one from which the previous player (i.e., the opponent) can win. We monitored the accuracy of the value network by assessing whether it outputs a positive value for winning positions and a negative value for losing positions. For 5-heap Nim, all possible board positions arising from the initial state were evaluated. However, due to the exponentially growing state spaces of 6-heap and 7-heap Nim, evaluations for these variants were conducted on 10,000 randomly sampled board positions. As shown in Figure \ref{fig:alpha_value_accuracy}, the value network's performance significantly degrades with increasing heap count (and thus state space size and board complexity). The value network's predictive accuracy for 7-heap Nim barely surpassed random chance (50\% accuracy), demonstrating a severe breakdown in its ability to reliably evaluate game states.

\begin{figure}[ht]
\centering
\includegraphics[width=0.48\textwidth]{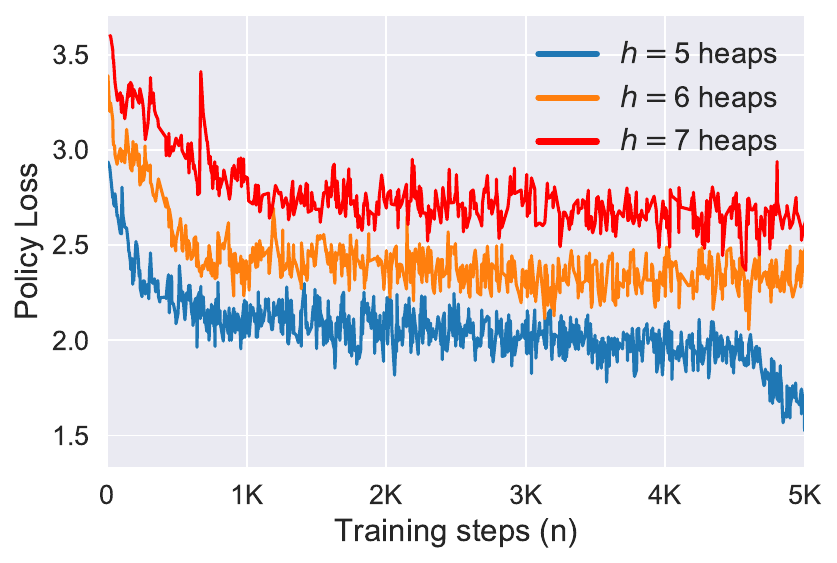}
\hspace{0.2em}
\includegraphics[width=0.48\textwidth]{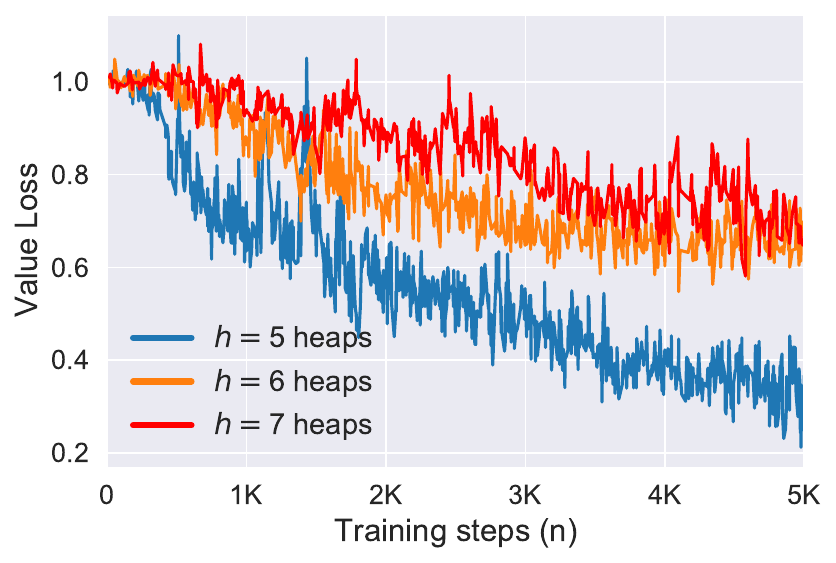}
\caption{\label{fig:alpha_value} Training loss of the policy network (left) and the value network (right) for Nim of 5, 6, and 7 heaps. For both neural networks, fitting the MCTS-derived heuristics and game outcomes becomes increasingly challenging as the board size grows.}
\end{figure}

The policy network learns from the search probabilities $\boldsymbol\pi(\textbf{a}|s)$ derived from the MCTS, as detailed in Formula \ref{eqn:pi_action_selection}. The value network learns from the actual game outcomes recorded during self-play. To further analyse the learning dynamics of these components, we tracked their respective training losses throughout the process, presented in Figure \ref{fig:alpha_value}. It is evident that both policy and value networks can progressively fit their respective targets, which contributes to the agent's increasing competitiveness (as reflected by Elo). However, this convergence in loss does not translate into expert-level accuracy. The inherent difficulty for the policy network to accurately learn the complex heuristics and for the value network to precisely model expected outcomes intensifies with increasing board size. This foundational problem significantly impedes the agent's progression towards true expert-level play.

The consistently decreasing training losses of the PV-NN, coupled with the rising Elo ratings of the agents, strongly suggest that the agent is learning to become a 'champion' within its self-play ecosystem. However, the concurrent and significant decline in the PV-NN's accuracy as the board size increases underscores a fundamental challenge in achieving 'expert' mastery. On 7-heap Nim, the PV-NN appears capable of memorising the heuristics and game outcomes from frequently encountered self-play data. Crucially, this rote memorisation is insufficient to provide the robust, generalizable guidance needed by the MCTS to explore the vast state space effectively and form a truly constructive policy improvement loop. This highlights a critical disconnect where fitting training data does not equate to learning the underlying mathematical principles of the game.

\subsubsection{Analysis on representative Nim positions}
\label{subsec:impartial_analysis}

While the preceding sections provided a high-level performance overview across various Nim board sizes, this section delves into specific case studies to provide a fine-grained analysis of the algorithm's behaviour. We evaluate the performance of our trained AlphaZero agent on selected initial and intermediate positions for 5, 6, and 7-heap Nim. This approach aligns with methodologies used in analysing game AI, such as the statistical analysis of LCZero's performance on particular Chess positions in Section \ref{sec:impartial_revisiting}.

\begin{figure}[htbp]
    \centering
    \subfloat[\label{pos:5heaps_1}{5 heaps: [1, 3, 5, 7, 9]}]{\includegraphics[width=0.29\textwidth]{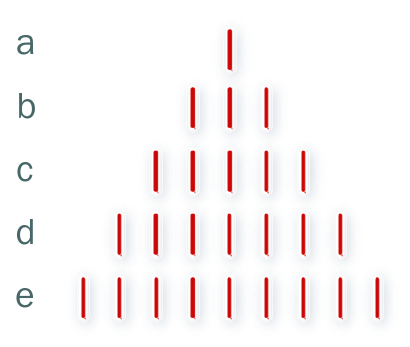}}
    \centering
    \hspace{1em}
    \subfloat[\label{tab:heaps5_1}Evaluations from policy and value network for the 2 moves with the highest prior probabilities. The win probability reflects the empirical win rate from MCTS simulations. In 10,000 MCTS simulations, only these two moves were significantly explored on this position.]{
        \begin{tabular}{p{0.3\textwidth}p{0.1\textwidth}p{0.1\textwidth}}\hline
            \toprule
            \textbf{Move} & e9 & a1 \\
            \midrule
            \textbf{Winning Move} & yes &  no \\
            \midrule
            \textbf{Prior Probability} & 97.9\% & 1.9\% \\
            \midrule
            \textbf{Win Probability} & 99.5\% & 5.0\%  \\
            \midrule
            \textbf{V-value} & 0.97 & -0.89  \\
            \bottomrule
        \end{tabular}
        \vspace{0.1em}
    }
    \caption{\label{fig:alphazero_5piles_1} On the initial position of 5-heap Nim, the policy network accurately assigns a 97.9\% prior probability to the winning move 'e9', with MCTS simulations yielding a 99.5\% empirical win probability. The move with the second highest prior probability 'a1' (a losing move) is assigned a 5.0\% winning probability.}
\end{figure}

The analysis of the initial board position for a 5-heap Nim game, as evaluated by our trained PV-NN, is presented in Figure \ref{fig:alphazero_5piles_1}. All values are computed from the perspective of the current player. Moves are denoted by a heap label (e.g., 'e') followed by the number of counters removed from that heap (e.g., 'e9' signifies removing 9 counters from heap 'e'). In this specific initial position, the policy network assigns an exceptionally high prior probability of 97.9\% to the winning move (e9), and the value network accurately estimates the resulting state's advantage for the current player. While this strong, accurate prioritisation of the optimal move is highly desirable for efficient search, it also underscores a potential vulnerability: an ungrounded overconfidence in an erroneous prior could lead to catastrophic and irrecoverable search misdirections, where MCTS simulations are unable to correct the initial bias.

\begin{figure}[htbp]
    \centering
    \subfloat[\label{pos:5heaps_2}{5 heaps: [1, 3, 5, 5, 9]}]{\includegraphics[width=0.3\textwidth]{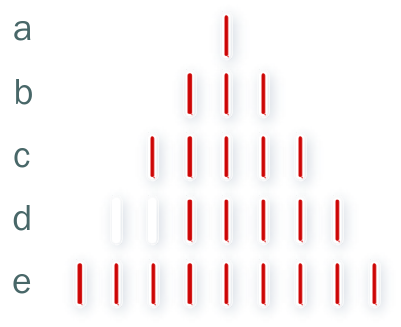}}
    \centering
    \subfloat[Evaluations from policy and value network for the 4 moves with the highest prior probabilities.]{
        \begin{tabular}{p{0.25\textwidth}p{0.08\textwidth}p{0.06\textwidth}p{0.06\textwidth}p{0.06\textwidth}}\hline
            \toprule
            \textbf{Move} & e8 & b1 & a1 & d4 \\
            \midrule
            \textbf{Winning Move} & no &  no & no &  no \\
            \midrule
            \textbf{Prior Probability} & 97.4\% & 0.7\% & 0.4\% & 0.2\%\\
            \midrule
            \textbf{Win Probability} & 0.14\% & 5.6\% & 12.5\% & 9.8\% \\
            \midrule
            \textbf{V-value} & -0.99 & -0.88 & -0.74 & -0.80 \\
            \bottomrule
        \end{tabular}
    }
    \vspace{0.1em}
    \subfloat[\label{tab:heaps5_2}Posterior probability of the true winning move 'e7' after different numbers of MCTS simulations.]{
        \begin{tabular}{cccccc}
            \toprule
             & \multicolumn{5}{c}{\textbf{Number of simulations}} \\
            \midrule
            \textbf{Winning Move}&  64 & 256 & 1024 & 65536 & 4194304 \\
            \midrule
            e7 & 1.56\% & 0.39\% & 0.09\% & 0.0015\% & 0.00021\% \\
            \bottomrule
        \end{tabular}
    }
    \caption{\label{fig:alphazero_5piles_2} For this position, the unique winning move is 'e7'. However, the policy network assigns an overwhelming 97.4\% prior probability to 'e8', a losing move. Despite the value network accurately predicting the resultant position of taking 'e8' is disadvantageous for the current player, the algorithm fails to find the winning move 'e7' after more than 4 million simulations.}
\end{figure}

In contrast, Figure \ref{fig:alphazero_5piles_2} illustrates an intermediate 5-heap Nim position where the MCTS is misled. The policy network exhibits an extreme overconfidence in the losing move 'e8', assigning it an overwhelming prior probability of 97.4\%. Crucially, the true winning move for this position is 'e7', which is not among the top-prioritised moves by the policy network. Despite the value network correctly predicting that taking 'e8' leads to a disadvantageous position for the current player (V-value of -0.99), the algorithm entirely fails to identify the winning move 'e7' even after an extensive 4 million MCTS simulations. The posterior probability of 'e7' actually decreases with more simulations (Table \ref{tab:heaps5_2}), indicating the MCTS is reinforcing the incorrect prior rather than correcting it. This failure is even more pronounced than the challenges observed in certain Tesuji problems in Go, where millions of simulations might eventually yield the best action \citep{shih2022novel}.

\begin{figure}[htbp]
    \centering
    \subfloat[\label{pos:6heaps_1}{6 heaps: [1,3,5,7,9,11]}]{\includegraphics[width=0.3\textwidth]{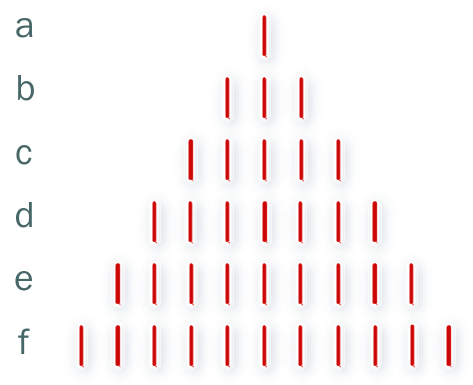}}
    \centering
    \subfloat[\label{tab:heaps6_1}Evaluations from policy and value network for the 4 moves with the highest prior probabilities.]{
        \begin{tabular}{lcccc}\hline
            \toprule
            \textbf{Move} & b2 & a1 & b1 &  d6 \\
            \midrule
            \textbf{Winning Move} & yes &  no & no &  no \\
            \midrule
            \textbf{Prior Probability} & 92.8\% & 4.0\% & 0.64\% & 0.63\%  \\
            \midrule
            \textbf{Win Probability} & 78.0\% & 14.3\% & 4.76\% & 2.17\%  \\
            \midrule
            \textbf{V-value} & 0.56 & -0.71 & -0.90 & -0.95  \\
            \bottomrule
        \end{tabular}
        \vspace{0.01em}
    }
    \caption{\label{fig:alphazero_6piles_1} For the initial 6-heap Nim position, the policy and value networks accurately predict the winning move 'b2' and estimate its resulting state value. The prior probability assigned to 'b2' significantly surpasses that of other moves.}
\end{figure}

Turning to the 6-heap Nim, Figure \ref{fig:alphazero_6piles_1} shows the analysis for its initial position. Here, the policy network strongly favours the winning move 'b2', and the value network confidently and accurately evaluates the resulting positions. This indicates that for the initial, frequently encountered states of moderately sized boards, the PV-NN can still provide effective guidance.

\begin{figure}[H]
    \centering
    \subfloat[\label{pos:6heaps_2}{6 heaps: [1,3,3,5,4,10]}]{\includegraphics[width=0.3\textwidth]{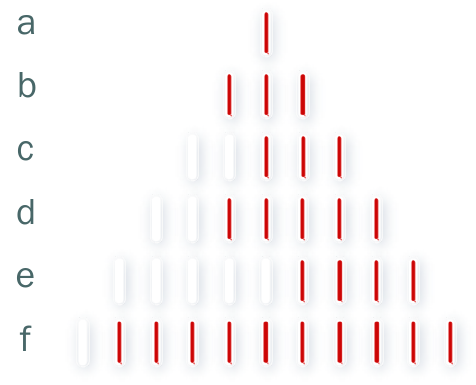}}
    \subfloat[Evaluations from policy and value network for the 4 moves with the highest prior probabilities.]{
        \begin{tabular}{lcccc}\hline
            \toprule
            \textbf{Move} & f3 & f2 & f4 &  f9 \\
            \midrule
            \textbf{Winning Move} & no & no & no &  no \\
            \midrule
            \textbf{Prior Probability} & 47.4\% & 12.6\% & 10.1\% & 7.5\%  \\
            \midrule
            \textbf{Win Probability} & 0.18\% & 1.03\% & 0.03\% & 81.2\%  \\
            \midrule
            \textbf{V Value} & -0.99 & -0.97 & -0.99 & 0.62  \\
            \bottomrule
        \end{tabular}
    }
    \vspace{0.1em}
    \subfloat[\label{tab:heaps6_2}Posterior probability of the true winning move 'f10' given different numbers of MCTS simulations.]{
        \begin{tabular}{cccccc}
            \toprule
             & \multicolumn{5}{c}{\textbf{Number of simulations}} \\
            \midrule
            \textbf{Winning Move}&  64 & 256 & 1024 & 65536 & 4194304 \\
            \midrule
            f10 & 4.68\% & 1.17\% & 0.29\% & 53.7\% & 99.1\% \\
            \bottomrule
        \end{tabular}
    }
    \caption{\label{fig:alphazero_6piles_2} In this position, the only winning move is 'f10'. The policy network's top 4 prior probabilities are assigned to losing moves, initially hindering the search. While the value network sometimes mispredicts the outcome of certain losing moves (e.g., 'f9' having a high empirical win probability), the MCTS eventually identifies and strongly favours the winning move 'f10' after 65,536 simulations, and its confidence is further boosted with more simulations.}
\end{figure}

However, similar to 5-heap Nim, there are also challenging 6-heap positions where the PV-NN falters. One such example is shown in Figure \ref{fig:alphazero_6piles_2}. Here, the top four prior probabilities from the policy network are incorrectly assigned to losing moves (Table \ref{tab:heaps6_2}). Unlike the severe misdirection seen in Figure \ref{fig:alphazero_5piles_2}, the policy network's confidence in these incorrect moves is not overwhelmingly high (e.g., 47.4\% for 'f3' vs. 97.4\% for 'e8' previously), allowing the MCTS to explore alternative branches more effectively. Although the initial MCTS empirical win probability for some losing moves (e.g., 'f9') and their V-values are misleadingly high, indicating the value network is still prone to mis-evaluating certain losing states, the MCTS eventually identifies the true winning move, 'f10', after 65536 simulations. Its posterior probability (Table \ref{tab:heaps6_2}, bottom) subsequently increases significantly with further simulations, demonstrating MCTS's compensatory power in this instance, despite initial poor policy priors and some inaccurate value evaluations.

\begin{figure}[ht]
    \centering
    \subfloat[\label{pos:7heaps}{7 heaps: \newline[1, 3, 5, 7, 9, 11, 13]}]{\includegraphics[width=0.3\textwidth]{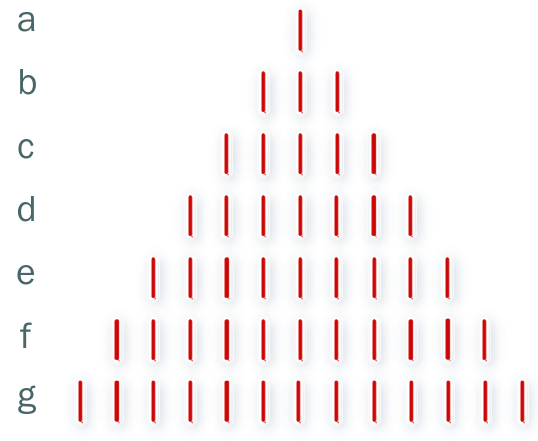}}
    \centering
    \subfloat[Evaluations from policy and value network for the 4 moves with the highest prior probabilities.]{
        \begin{tabular}{lcccc}\hline
            \toprule
            \textbf{Move} & c4 & c3 & c2 &  c5 \\
            \midrule
            \textbf{Winning Move} & no & no & no &  no \\
            \midrule
            \textbf{Prior Probability} & 4.37\% & 4.06\% & 3.95\% & 3.88\%  \\
            \midrule
            \textbf{Win Probability} & 50.1\% & 49.5\% & 48.9\% & 50.1\%  \\
            \midrule
            \textbf{V Value} & 0.003 & -0.009 & -0.02 & 0.003  \\
            \bottomrule
        \end{tabular}
    }
    \vspace{0.1em}
    \subfloat[Posterior probability of the true winning moves 'e7', 'f7', and 'g11' given different numbers of MCTS simulations.]{
        \begin{tabular}{cccccc}
            \toprule
             & \multicolumn{5}{c}{\textbf{Number of simulations}} \\
            \midrule
            \textbf{Winning Move}&  64 & 256 & 1024 & 65536 & 4194304 \\
            \midrule
            e7 & 1.56\% & 1.17\% & 1.26\%  & 2.79\% & 0.38\% \\
            \midrule
            f7 & 1.56\% & 0.17\% & 1.66\% & 0.89\% & 1.15\% \\
            \midrule
            g11 & 1.56\% & not visited & 1.07\% & 1.61\% & 0.65\%  \\
            \bottomrule
        \end{tabular}
    }
    \caption{\label{fig:alphazero_7piles} For this 7-heap Nim position, the policy network's prior probabilities are nearly uniform, with the top 4 assigned to losing moves. The value network's evaluations are near zero (approximating 50\% win probability). Despite extensive MCTS, the true winning moves ('e7', 'f7', 'g11') are not recognised, and their posterior probabilities remain negligibly low.}
\end{figure}

Finally, for the initial position of 7-heap Nim (Figure \ref{fig:alphazero_7piles}), which possesses three winning moves, the policy network assigns nearly uniform prior probabilities across all legal moves. Concurrently, the value network evaluates all resulting positions with near-zero values, effectively indicating an inability to reliably differentiate between winning and losing states (a 50\% winning probability for all moves). These non-informative predictions critically undermine the MCTS, as demonstrated by its failure to identify any of the winning moves, even after more than 4 million simulations. 

Across Nim variants with 5, 6, and 7 heaps, our analysis reveals a duality: positions where the PV-NN succeeds in evaluation alongside positions where it fundamentally fails. The high accuracy of predictions on initial positions of smaller boards suggests a greater learned knowledge for frequently encountered states. However, the confidence in prioritising optimal moves deteriorates, and the ability to accurately evaluate general positions escalates in difficulty as the board size and associated state space grow.

On relatively small boards, the underlying parity computations for nim-sum remain computationally feasible, and the neural networks can approximate this function to a reasonable degree. However, mastering Nim on larger boards becomes dramatically more challenging because identifying the optimal move and accurately evaluating large board positions critically entails solving complex parity-related issues using neural networks. The immense state space necessitates the PV-NNs to generalise effectively to unseen states, which, as previously argued, proves exceptionally difficult for functions with high complexity like parity. Furthermore, the inherent difficulties are compounded by the unknown and unpredictable data distribution generated by self-play reinforcement learning, as the characteristics of training data profoundly influence a neural network's capacity to model perfect parity functions \citep{daniely2020learning, cornacchia2023mathematical}. The persistent problem of noisy labels, a known challenge in self-play RL, further exacerbates the difficulty for the PV-NN to learn an accurate parity function \citep{zhou2023exploring}. Collectively, these complexities represent a fundamental and significant challenge to contemporary reinforcement learning algorithms in their pursuit of mastering Nim. Our results empirically demonstrate that in a real-world self-play RL setting, the parity function presents a more substantial obstacle for an RL agent to achieve expert-level proficiency in Nim. Given that all impartial games can be theoretically translated into an equivalent Nim game, these identified challenges naturally extend to the broader class of impartial games.

\subsection{Controlled experiments}

The findings in Section~\ref{sec:impartial_experiments} raise the question of whether the challenges of AlphaZero-style RL on impartial Nim are fundamentally caused by limitations of reinforcement learning (e.g., insufficient exploration, optimisation instability, data sample out-of-distribution during training), or instead by the difficulty neural networks face in learning the parity-based winning strategy required for optimal play. To disentangle these factors, we design two controlled experiments on small game boards where the entire state space is tractable and repeatedly revisited, thereby removing exploration difficulty, function underfitting due to limited data, and instability in value backpropagation as confounding variables.

The first experiment investigates whether the difficulty lies specifically in neural network function approximation. We replace the neural policy and value networks with tabular lookup tables that can represent the optimal mapping exactly without any inductive bias assumptions. If the tabular tables learn successfully while the neural networks struggle, this would isolate the challenge to neural network generalisation, rather than RL dynamics or game complexity.

The second experiment examines whether the difficulty is intrinsic to the Nim strategy itself. We construct a structurally similar partisan variant of Nim whose optimal strategy does not rely on learning a parity function. If AlphaZero with PV-NNs succeeds in this modified game under identical training conditions, this would further indicate that the challenge in classical Nim originates from the necessity of learning a non-associative, non-linear arithmetic function, not from our modelling choices, optimisation procedure, hyperparameters, or representation.

\subsubsection{Tabular AlphaZero-style learning on impartial Nim}

To investigate whether the generalisation difficulties of neural networks in learning parity contribute to the challenges faced by AlphaZero-style reinforcement learning on Nim, we implemented a tabular variant of the algorithm where we replaced the PV-NNs with two tabular lookup tables representing the policy and value functions, respectively. This tabular variant serves as a baseline that circumvents neural network approximation errors, allowing us to isolate the impact of function approximation on learning and generalisation in the parity-based Nim environment.

For each state $s$ in the state space, the policy table is initialised to assign equal probability to all legal actions, while the value table is initialised to zero. Given a target policy $\boldsymbol\pi_{\text{target}}(\mathbf{a}|s)$ for a state $s$ obtained from MCTS, the policy table is updated using a moving average:

\begin{equation}
\boldsymbol\pi(\mathbf{a}|s) \leftarrow (1 - \eta)\,\boldsymbol\pi(\mathbf{a}|s) + \eta\,\boldsymbol\pi_{\text{target}}(\mathbf{a}|s),
\end{equation}

where $\eta \in (0,1)$ is a learning rate controlling the update magnitude. After each update, the policy is normalised to ensure it forms a valid probability distribution. Similarly, given a target value $v_{\text{target}}(s)$ from MCTS, the value table is updated via

\begin{equation}
v(s) \leftarrow \text{clip}\big((1 - \eta)\,v(s) + \eta\,v_{\text{target}}(s), -1, 1\big),
\end{equation}

where $\text{clip}(\cdot,-1,1)$ ensures the value remains within range as used in AlphaZero. 

\begin{figure}[H]
\centering
\includegraphics[width=0.45\textwidth]{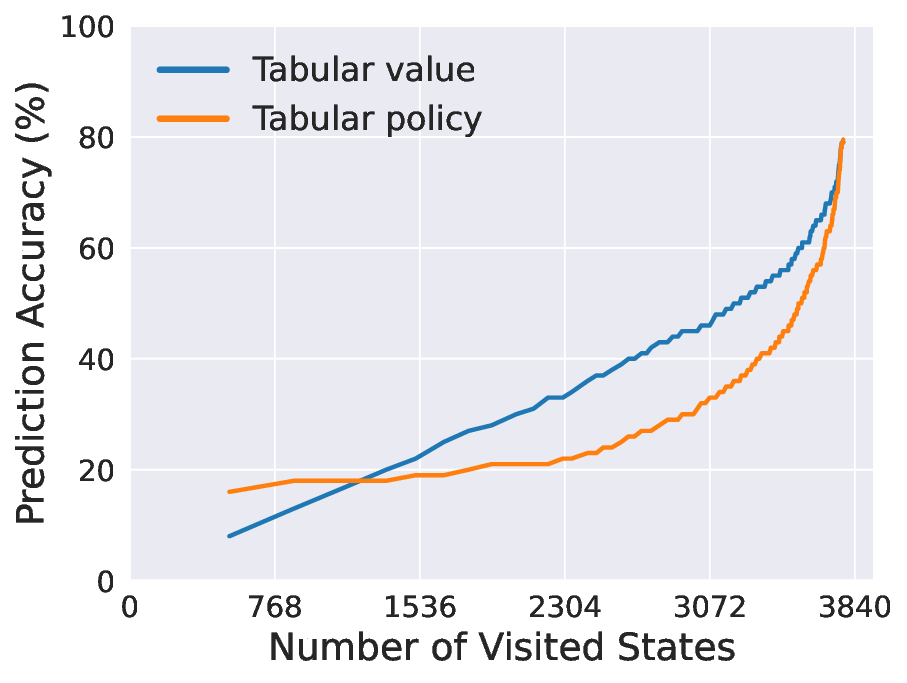}
\caption{\label{fig:tabular_pv} Prediction accuracy of the tabular policy and value models versus the number of distinct states explored during the Tabular AlphaZero-style RL training on the five-heap impartial Nim.}
\end{figure}

We applied the Tabular AlphaZero-style algorithm to the five-heap Nim game while keeping all algorithmic and training configurations identical to those used in Section~\ref{sec:impartial_experiments}, ensuring a strictly controlled comparison. Figure~\ref{fig:tabular_pv} reports the learning progress, measured by the policy and value prediction accuracy as a function of the number of distinct states visited across 500 training iterations. Both the tabular policy and value estimates improve steadily as more states are encountered, eventually achieving $80$\% accuracy on the states in the state space. Notably, the tabular policy accuracy is approximately twice that achieved by the policy network (see Fig.~\ref{fig:alphazero_policy_accuracy}(a)). This demonstrates that, on this small Nim board, the challenge does not stem from the AlphaZero training framework itself, but rather from the difficulties of neural networks in learning parities.

Interestingly, the gap between the tabular lookup table and the value network is substantially smaller for value estimation. The final value accuracy of the tabular value closely matches that of the corresponding value network (Fig.~\ref{fig:alpha_value}). This suggests that value prediction, a scalar regression task, is considerably easier to generalise across unseen states than policy prediction, which requires learning a high-dimensional, parity-structured mapping over combinatorial action spaces. Taken together, these results indicate that, within this small five-heap Nim setting, the problem is not due to insufficient exploration, but arises from the difficulty neural networks have in representing the parity-based structure required for accurate policy generalisation.

\subsubsection{AlphaZero-style learning on Partisan Nim}

To examine whether the learning difficulty observed in mastering Nim primarily arises from the parity issue, we introduce a partisan variant of the game in which each player controls a disjoint subset of heaps and is only allowed to remove matches from those heaps. A player loses if no matches are available on the set of heaps they can remove matches from. Unlike classical Nim, this partisan setting eliminates the need for parity reasoning. Each player’s winning potential depends solely on the total number of matches available within their own designated heaps. As a result, the optimal policy becomes strictly additive and memoryless. Both players should remove exactly one match per turn, thereby maximising the number of future turns they can take. Under optimal play, the winner is the player with the larger initial sum of matches.

We evaluated this setting using the same initial board configuration 
on five heaps $\,[1,3,5,7,9]\,$ as in our impartial Nim experiments. The only modification lies in the move constraints: Player~1 can remove matches exclusively from heaps labelled $a$, $c$, and $e$, while Player~2 is restricted to heaps $b$ and $d$. All algorithmic and representation settings are kept identical to the baseline experiments, allowing us to isolate the effect of removing parity-based reasoning from the learning problem.

Within ten self-play training iterations, the policy network rapidly converges to the theoretically optimal strategy. The trained policy network assigns its highest action probability to removing exactly one match from any legal heap, regardless of which player is to move. For example, in a mid-game position 
$\,[1,3,5,5,2]\,$ where both players have the same number of matches, the best move for each player is to remove one match from one of the designated matches. When this position is labelled as Player~1’s turn, the policy network assigns probabilities of $0.53$, $0.29$, and $0.11$ to removing one match from heaps $e$, $c$, and $a$, respectively. When labelled as Player~2’s turn, it assigns $0.64$ and $0.19$ to actions $b1$ and $d1$, respectively, again matching the optimal strategy.

This stark contrast between the agent's performance on impartial Nim (which struggled significantly) and its quick mastery of partisan Nim supports our central finding that the inherent difficulty for generic neural networks to implicitly model abstract, non-associative functions like parity is the primary bottleneck in this learning paradigm. Furthermore, its exceptional performance using the same board representation as in the impartial Nim experiments suggests that the game state representation itself does not adversely impact performance when the underlying optimal function is simpler and more associative.

\subsection{Limitations and future work}

We acknowledge that the re-implementation of a method and its subsequent underperformance on a new domain does not inherently imply a flaw in the original method itself. A comprehensive empirical exploration would typically involve an exhaustive ablation study across a vast hyperparameter space, including neural network architectures, representations, and training schedules. However, our primary aim is not to definitively prove an empirical impossibility for AlphaZero-style algorithms to master Nim. Instead, our objective is to precisely diagnose and illustrate the fundamental challenges and inherent difficulties these algorithms encounter when confronted with games whose optimal strategy relies on abstract mathematical properties rather than associative pattern recognition.

To contextualise this, consider our analysis of a specific Chess position (Fig. \ref{fig:LC2}) which subtly mirrors a rudimentary parity (nim-like) configuration. Our investigation revealed that even the highly optimised and extensively trained LCZero, which for all practical purposes embodies the AlphaZero paradigm, struggled to accurately assess this particular position. This observation provides a compelling precursor to our central argument: in Nim, where the winning strategy is directly and inextricably linked to the parity function (nim-sum), this issue becomes dramatically more pronounced. The core difficulty lies in the generic neural network's struggle to implicitly learn a perfect parity function from noisy, self-play generated data, especially as the input dimensionality increases. This inherent representational bottleneck for combinatorial logic, distinct from the complexity of typical board game patterns, is what we aim to thoroughly explore in this work.

Our current approach employs an LSTM network over a structured representation of Nim states. While this provides a simple and flexible sequence-learning interface that can accommodate inputs of arbitrary length, which is consistent with the nature of the parity function, it does not explicitly capture the combinatorial structure or the permutation invariance inherent to impartial games. In Nim, permuting heap order does not alter optimal play, yet our representation and LSTM implicitly introduce order sensitivity, yielding a potential inductive bias mismatch between architecture and game semantics. More structure-aware neural networks, such as permutation-invariant encoders (e.g., Deep Sets \citep{zaheer2017deep}), or Transformers without positional encodings \citep{haviv2022transformer} may be better aligned with the symmetry and parity properties of impartial games, and could improve sample efficiency and generalisation. As we have surveyed in Section~\ref{sec:learning_parity_nn}, these established neural architectures still face intrinsic challenges in modelling a perfect parity function, especially from long binary strings. We are therefore interested in evaluating alternative neural architectures as PV-NNs for learning impartial games, and we leave this to future work.

Although we adopt the standard AlphaZero framework, our study does not evaluate several notable extensions that could potentially strengthen performance. For example, Gumbel AlphaZero \citep{danihelka2022policy} has demonstrated improved exploration and policy improvement guarantees, and recent AlphaZero-derived solving frameworks \citep{wu2022alphazero, wu2023game, dantsin2023alphazero} explicitly target exact reasoning over full game graphs. Exploring such variants, particularly in settings where theoretical optimal play is known and parity-based reasoning is central, represents an important direction for future investigation. A more comprehensive empirical comparison across these methods may yield deeper insight into the challenges identified in this work. We leave this for future work.

\section{Concluding remarks and conjectures}
\label{sec:impartial_concluding}

The AlphaZero-style learning paradigm represents a monumental achievement in the quest for creating intelligent systems that can learn to solve complex tasks from first principles. Its success in mastering games like Chess, Shogi, and Go stands as a powerful testament to the efficacy of combining deep neural networks with a self-play-driven search algorithm. However, as articulated by Demis Hassabis, the ambition to "solve intelligence" and apply it universally hinges on an agent's ability to handle diverse modes of thinking. The empirical results presented in this paper reveal a fundamental challenge to this ambition: the AlphaZero paradigm, in its current form, is profoundly limited by its inductive biases and representational capacity, which are ill-suited for games requiring abstract, combinatorial reasoning.

From a cognitive science perspective, the difference between games like Chess (or Go) and Nim is not merely one of complexity but of the very nature of optimal strategy. Mastering Chess and Go primarily involves associative pattern matching, heuristic evaluation of board configurations, and deep subsymbolic calculation. A human Grandmaster's mastery is built upon millions of learned patterns, making it an ideal domain for reinforcement learning agents that excel at function approximation on large, but structured, data distributions. In contrast, Nim is governed by a single, elegant mathematical principle: the nim-sum. A human's mastery of Nim is non-incremental; it is achieved not through extensive pattern recognition but through a single, abstract insight into combinatorial game theory. This disparity highlights a crucial distinction: AlphaZero excels at learning through association, but fails when a problem requires a form of symbolic reasoning that cannot be implicitly learned from the correlation between game states and outcomes.

Our work empirically demonstrates that this theoretical vulnerability becomes a tangible, catastrophic failure mode in practice. While the problem of a neural network's difficulty in learning parity from a uniform distribution is a well-established theoretical finding, our contribution is to show how this representational bottleneck manifests and breaks the AlphaZero learning loop in a real-world, self-play setting. The motivation behind our research is not to simply identify a known weakness, but to meticulously dissect its consequences and chart a path for extending the AlphaZero paradigm. By analysing the breakdown of the policy and value networks on specific Nim positions, we have revealed how a critical representational gap can trigger a cascade of failures.

The core of the problem, as our analysis shows, is a breakdown in the crucial positive feedback loop that underpins reinforcement learning. The agent's performance on small boards, while seemingly successful, is based on a precarious "champion" notion of mastery, memorising a few optimal opening moves. This allows the agent to steer the game into a sub-space it has seen frequently. However, as the number of heaps increases to just seven, the policy network's predictions become near-uniform, and the value network's assessments drift to a coin-toss probability. This lack of a strong initial policy prevents the MCTS from exploring the state space effectively, leaving it unable to find winning moves, even with millions of simulations. This is not a failure of calculation; it is a failure of learned guidance. Crucially, this happens even on boards where the state space is still small enough that many states would be revisited during training. In this scenario, the learning process becomes completely derailed, as the agent fails to generalise to any new positions, regardless of whether the parity function is complex. When the state space becomes truly vast (e.g., with 50+ heaps), the difficulty of learning the abstract parity function for large numbers exacerbates an already fatal condition. The network's inability to generalise to unseen positions adds profound noise to the learning process, which is already struggling to handle the non-associative nature of the game, creating a vicious cycle that makes true expert-level mastery impossible.

Our findings lead to several key conjectures and open questions for the machine learning community:
\begin{enumerate}
\item \textbf{The Need for Neuro-Symbolic Hybrid Architectures}: The AlphaZero architecture appears to be a representational bottleneck for problems that require both pattern matching and symbolic reasoning. We conjecture that a more powerful paradigm will emerge from hybrid systems that combine a neural network for its strengths in subsymbolic pattern recognition with a separate symbolic reasoning module explicitly designed to handle combinatorial logic (e.g., parity, XOR operations, or other abstract game principles). This would allow the system to learn from data while also having the capacity to derive and apply abstract rules.
\item \textbf{Expanding the Action Space with Meta-Actions}: In games like Nim, a human's path to mastery is not a sequence of moves but a process of abstract thought: they might "scribble on paper" to calculate nim-sums or "form a conjecture" about a winning strategy. We propose a future paradigm where the agent's action space is expanded to include such "meta-actions" that are external to the game's move mechanics. These meta-actions could involve querying an external symbolic memory, performing a specific logical operation, or invoking a pre-programmed solver for a sub-problem. This would be analogous to an AlphaZero-style agent having the ability to perform a strategic calculation, which is not a move itself but dramatically boosts its ability to plan forward. The challenge, as our results indicate, is to design an architecture that can learn when and how to invoke these meta-actions effectively.
\item \textbf{The Champion-Expert Dichotomy as a New Evaluation Metric}: Our work highlights that winning a game and truly understanding it are not the same. We propose that the distinction between a "champion" (an agent that wins from the start state) and an "expert" (an agent that plays optimally from all states) is a crucial and under-explored metric for evaluating the true mastery and generalisation capabilities of reinforcement learning algorithms. This distinction is especially critical for domains where adversarial attacks or rare, but strategically important, positions can expose the limitations of an agent's learned knowledge.
\end{enumerate}
In conclusion, while AlphaZero has provided a powerful template for solving complex games, its current form possesses a profound and fundamental limitation when confronted with problems that necessitate a blend of associative pattern recognition and abstract symbolic reasoning. Our work on impartial games provides a clear, empirical case study of this limitation, offering both a diagnosis of the failure modes and a set of concrete, thought-provoking conjectures for how the next generation of reinforcement learning algorithms might overcome this critical hurdle on the path to more intelligent machine learning agents.

\section*{Acknowledgements}

We appreciate the assistance of the IT Research team at Queen Mary University of London for supporting us in using the Apocrita HPC facility. We would also like to thank the LCZero development team for providing detailed instructions for the use and explaining the inner workings of LCZero.

\bibliographystyle{plainnat}
\bibliography{sample}

\end{document}